**Statement**

*This page provides supplementary information for the reader and is not part of the main article body or pagination.*

---

### 1. Language and Version Notice

This document is an **English translation of the original Chinese version** of this article.

- This English version is provided **for reference and reading convenience only** and **is not intended for formal publication**.
- In case of any ambiguity or discrepancy in meaning, the **original Chinese version shall prevail**.
- This English version may not be used as the final, citable, or archival version without explicit permission.

---

### 2. How to Download This Article

To download this article, along with any related data or supplementary materials, please follow the steps below:

1. Click the link below, or navigate to the specified address, and locate the article from the file list.

**Download / Access Link:**

🔗 [Insert link here, e.g., https://crad.ict.ac.cn/article/doi/10.7544/issn1000-1239.202550223]

*Note: If the above links do not work, please search for the article title or DOI (if available) on the "Journal of Computer Reasherch and Development" page or platform.*

---

### 3. How to Cite This Article

You are welcome to cite this article. Please choose **the original Chinese version**.

**Suggested Citation Format (fill in actual information):**

---

**Last updated:** 2026.05.19

**Contact (if you have questions or need permission): liztchina@hotmail.com**

# AI-Model Network: Concept, Current State and Future

Li Zhetao, Zeng Xiyu, Wang Jianhui, Xiao Yong, Liu Zhongren, Wu Junru, Lai Junjie, Huang Jijun, and Long Saiqin
（*College of information science and technology*, *Jinan University*, *Guangzhou* 510632）

**Abstract** While the primary function of computers lies in computation and processing, the core value of the Internet is rooted in sharing and collaboration. Computers create the Internet, and the Internet empowers the value of computers. The rapid development of the Internet, cloud computing, and big data is pushing artificial intelligence into the era of large models (LMs). However, the practical application of LMs is currently hindered by high training costs and deployment complexities, driving a shift toward lightweight, private, and domain-specific models. With the rapid proliferation and wide distribution of heterogeneous models, enabling effective interaction and collaboration among them has emerged as a critical bottleneck that urgently needs to be addressed in LM development. Drawing inspiration from the development of the Internet, this paper proposes the concept, vision, and system architecture of world wide AI-model network (AI-ModelNet). It is a novel paradigm that achieves interconnection, capability sharing, and collaborative reasoning by establishing pathways between models. We first briefly review the current state of single-model and multi-model research. Subsequently, the systemic vision and hierarchical architecture of AI-ModelNet are articulated, followed by validation of the framework’s feasibility through a prototype system and diverse application cases. Finally, key directions for future research are discussed preliminarily.

The value of computers lies in computation and processing, while the value of the internet lies in sharing and collaboration. Throughout the evolution of computing, widespread technological advancement has driven a pivotal shift in research focus—optimizing standalone machines toward architecting networked communication systems. In 1974, the advent of the TCP/IP protocol established a standard for network packets transmission, laying the foundation for the modern internet. In 1989, the HTTP protocol fueled the rise of the WWW [1], fundamentally transforming the way to exchange information, which has turned digital space, information space and network space into vital conduits for information sharing. Therefore, computing gave birth to the Internet, the Internet reciprocally drives the continual expansion of computing's application frontiers. They reinforce each other, with interconnection magnifying computing's core value.

This pattern of evolution from simple to complex is mirrored in the trajectory of artificial intelligence (AI). The evolution of AI models has broadly progressed through three distinct phases: the neural network phase, the deep learning phase, and the large models phase. During the neural network phase, models featured relatively simple architectures and focused primarily on solving basic classification and regression problems. This phase laid the essential groundwork for the subsequent deep learning revolution. In the deep learning phase, the progressive deepening of network architectures enabled models to capture more complex data features and meet increasingly demanding task requirements. These models saw widespread adoption in domains such as image recognition, speech recognition, and natural language processing. The introduction of the Transformer architecture [2] marked a pivotal transition into the era of large-scale pre-trained models, achieving substantial performance breakthroughs while dramatically improving parallel computing efficiency. As illustrated in Figure 1, computing evolved from simple operations to complex processing, while the

**收稿日期**：；**修回日期**：
**基金项目**：国家自然科学基金国际合作与交流重点项目(W2411053)；国家自然科学基金联合基金重点项目(U23B2027)
This work was supported by the Key Program of Projects of International Cooperation and Exchanges NSFC (W2411053) and the Key Project of the Joint Funds of the National Natural Science Foundation of China (U23B2027).
**通信作者**：李哲涛（liztchina@hotmail.com）

Internet drove a paradigm shift from isolated computing to global interconnection. Inspired by the developmental trajectory of the Internet, model interconnection emerges as a promising pathway to holistically amplify model value at the system level.

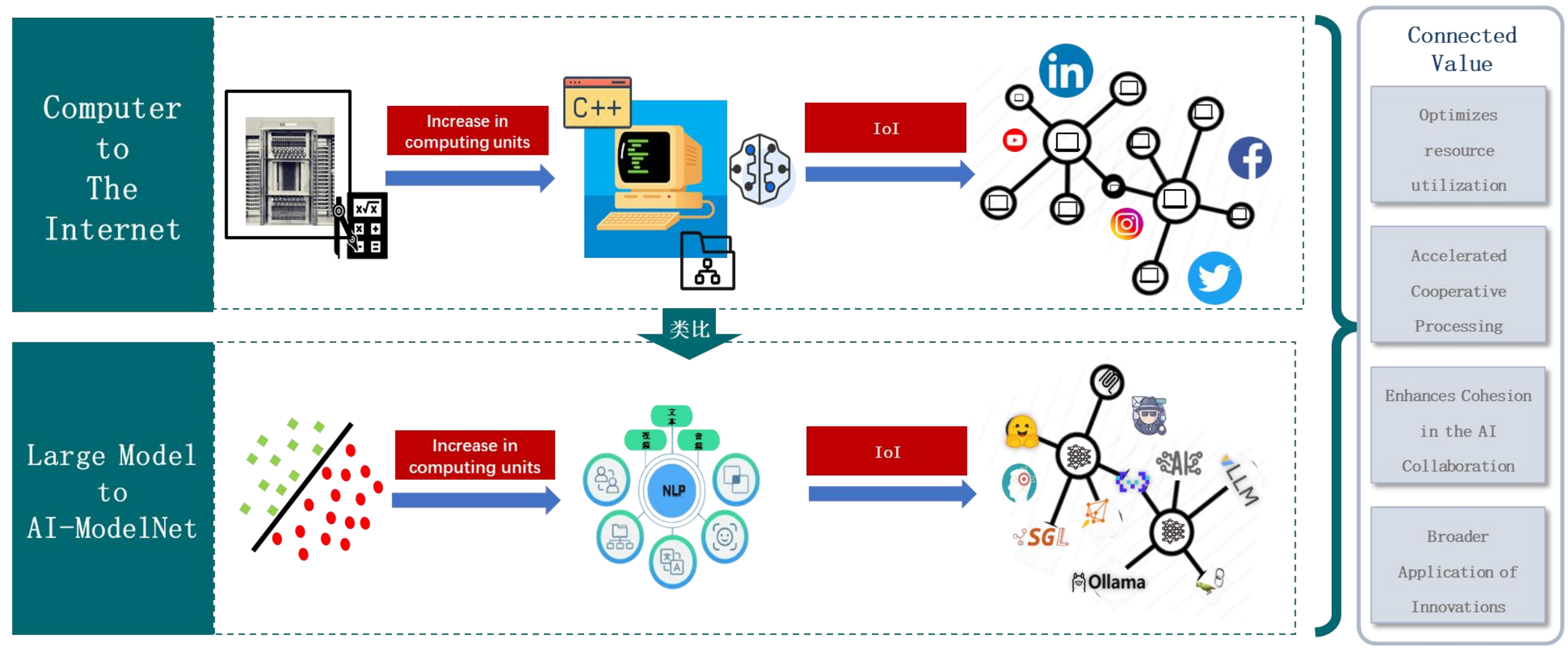

Fig. 1 From computer network to AI-ModelNet

The current era of AI is characterized by an explosion in parameters, fueled by large-scale data, vast computational resources, and the exponential growth of parameters and the corresponding demands for node-level compute pose tremendous challenges[3]. Consequently, lightweight, private, and vertical deployments have become mainstream trends. Currently, models are fragmented and lack interconnection, leading to the emergence of the “model silos” phenomenon. In practical applications, a single model faces multiple challenges, for instance, models can unintentionally absorb biases present in their training data[4], which affects the objectivity of their decisions; besides, when encountering knowledge gaps or uncertain scenarios, models are prone to hallucination, generating inaccurate outputs that can be influenced by their pre-existing biases[5], leading to the proliferation of false content. General models finetuning, and updates are resource-intensive technical processes. Domain-specific fine-tuning of general-purpose models frequently results in a loss of general capabilities[6]. The interconnection of computers boosts system value through shared resources and collaborative processing. By analogy, enabling models to interconnect and interoperate may serve as a pivotal mechanism in the pursuit of artificial general intelligence (AGI).

Higher-order capabilities in complex systems typically arise not from amplifying a single entity, but from the collective interactions and synergies of multiple bounded agents [5-9]. This principle is ubiquitous across natural, social, and engineered systems: Although, there are insurmountable boundaries between individual capabilities, the collections between individuals enable complementary information sharing, cognitive error correction, and amplified performance. From this prespective, the system's overall intelligence hinges more on its organizational design and interaction protocols than on the peak performance of individual components. This systemic principle has been empirically validated and is commonly captured by the saying “Two heads are better than one”—which centers on the premise that suitably organized collective wisdom can transcend the cognitive limits of any single individual. The essence lies in the crowd gain achieved through the multi-model collaboration: the differences among individual agents—in terms of knowledge perspectives, reasoning paths, and decision preferences—allow the system to continuously correct local biases, compensate for cognitive blind spots, and achieve more robust overall judgments through interaction [10]. The value of model collaboration is not to eliminate the perceptual limitations of each individual “blind” model, but to construct a robust consensus through mutual validation across diverse perspectives, and to progressively approach the holistic truth of the “elephant” by dynamically integrating local perceptions. Thus, inter-agent connections constitute a prerequisite for advanced collective intelligence.

Introducing this systematic perspective into AI field reveals that the interconnection among models is not an accidental choice but an inherent necessity. No matter how extensively a single model is scaled or its performance optimized, its internal representations remain constrained by structural capacity and data quality. Therefore, the connections between models provide opportunities for cross-model knowledge complementarity and collaborative reasoning. By establishing stable linkages among models, multiple models can collaborate to form collective intelligence, thereby transcending the inherent limitations of individual models at the systemic level. It embodies the philosophical principles of "universal connections" and "quantitative change leading to qualitative change." Specifically, as the collaboration among multiple models deepens, it can trigger a qualitative leap in their cognitive capabilities, task generalization, and autonomous coordination. According to the statistics, there are over 1,300 AI large models globally at present, covering major knowledge domains [11]. Thus, effectively organizing and coordinating existing models to accomplish diverse tasks has become a core challenge in enhancing the overall efficiency of intelligent systems.

As shown in Figure 1, if models are considered as encapsulated intelligent resources in computers, the interaction process among models can be essentially analogized to information transmission between network nodes. Therefore, this paper introduces the concept of AI-ModelNet: a novel paradigm that establishes standardized information pathways between models to enable dynamic discovery, on-demand composition, and collaborative reasoning of model capabilities. AI-ModelNet emphasizes interconnectivity and network-level collaboration among model nodes. It abstracts models into continuously discoverable and composable capability nodes, supporting open, heterogeneous, and evolvable intelligent systems. This approach decouples large models from massive computational power and data resources, effectively reducing the strong dependency of large-scale model technologies on such resources. Therefore, AI-ModelNet is poised to become a catalyst for the future development of AI.

## 1 Research Status

This section sequentially discusses the research progress in both existing single model and multi-model collaboration domains. Through systematic analysis, it elucidates the fundamental differences between AI-ModelNet and existing research paradigms.

### 1.1 Single Model

With the rapid advancement of large-scale pre-training technologies, single large models have demonstrated significant advantages in tasks such as natural language processing (NLP) and information retrieval. Propelled by the evolution of the Transformer architecture, an explosion of high-performance generative large models has given rise to the ear of model proliferation. Representative models such as the GPT series [12], the LLaMA series [13], the DeepSeek series [14], the Qwen series [15], and the Kimi series [16] have expanded their applications to diverse scenarios including language generation, knowledge-based question answering, reasoning, and tool invocation. Their deployment extends to fields such as autonomous driving [17], intelligent assistants [18], and literary creation [19], among others.

In practical applications, single models faces a series of systematic limitations. First, the knowledge and value boundaries of a model are constrained by its training data. Second, models may exhibit "hallucination" phenomena, generating information that is logically coherent but factually incorrect [20]. Additionally, the reasoning process of large models relies on substantial computational power, storage, and energy consumption [21], limiting their deployment in resource-constrained scenarios. Finally, specific training or fine-tuning for often sacrifice their general performance as a trade-off.

The limitations of single large model architectures-particularly in generalization, efficiency, and flexibility-are amplified when addressing challenges such as diverse task requirements, adaptation to complex environments, and long-term system evolution. It has propelled researches to explore multi-model architectures with collaborative capabilities. Employing mechanisms like task decomposition, model complementarity, and flexible routing, these multi-model architectures strive for modularized capabilities, transparent reasoning, and flexible model updates; this approach points toward novel paradigms in AI development.

### 1.2 Multi-model Collaboration

multi-model collaboration has recently emerged as a pivotal approach to building more flexible and efficient intelligent systems. The core concept of multi-model collaboration lies in enabling distributed reasoning for complex tasks and enhancing system robustness. This is achieved through mechanisms such as capability complementarity among models, task decomposition, and structured routing. Existing studies on these mechanisms are generally categorized into three strategic approaches [22]: model fusion, output integration, and dynamic routing.

Model fusion constructs a unified model by integrating the parameters of multiple pre-trained models, aiming for structural synergy and performance compression. Ties-Merging [23] introduces a lightweight strategy to mitigate parameter sign conflicts and magnitude imbalance issues. To address the issues of knowledge interference and performance degradation on heterogeneous test data, the Twin-Merging[24] method decomposes expert model knowledge into shared and exclusive components. It then dynamically fuses these two knowledge types according to the input, effectively resolving these challenges.

Output integration focuses on achieving more robust reasoning by fusing outputs while preserving model independence. DeePEn [25] introduces a bidirectional vocabulary mapping mechanism to bridge the semantic representation spaces of heterogeneous models. Wang et al. [26] introduce a token-level parallel collaboration method that connects multiple model nodes via a communication mechanism, enabling efficient cross-device interactive reasoning. Wu et al. [27] design a mutual-evaluation sequential collaboration mechanism to facilitate autonomous answer optimization among models. Additionally, multi-source knowledge integration frameworks [28] demonstrate the advantages of multi-model collaboration in task division and knowledge synthesis. Wu et al. [29] develop a collaboration-oriented multi-model system that enhances overall performance through role-based capability complementarity among agents.

Dynamic routing mechanisms enhance the flexibility of multi-model systems by dynamically selecting the most suitable models to execute subtasks during inference, thereby balancing performance and cost. Ong et al. [30] propose RouteLLM, which guides the router to learn model selection decisions based on human preference data. BEST-Route [31] combines model and sampling strategies for dual-level selection. GraphRouter [32] employs graph structures to model semantic relationships between models and queries. RCR-Router [33] introduces a role-aware memory mechanism for precise context information transmission. TO-Router [34] validates an efficient selection mechanism in industrial deployment scenarios, leveraging query features and expert capability mappings.

While the aforementioned methods have made positive strides in multi-model architecture design, current research still faces three major challenges: first, collaboration task design lacks universality and remains dependent on specific scenario configurations; second, the interaction mechanisms between models are rigid, making it difficult to adapt to the dynamic multi-task demands of real-world environments; third, the system heavily relies on centralized computational resources, resulting in high deployment costs and insufficient scalability. Although existing multi-model collaboration research has laid a solid foundation, its applicability in open, heterogeneous, and distributed environments remains to be further improved. This further highlights the critical value of "model interconnection" as a general infrastructure—by establishing a loosely coupled, communicable, and collaborative global model interconnection network, it can break through the boundaries of current collaboration paradigms and drive intelligent systems toward greater flexibility, scalability, and long-term autonomy.

In summary, while both multi-model collaboration systems and AI-ModelNet aim to achieve model cooperation and capability complementarity, they are built upon distinct paradigmatic approaches. The former often relies on static architectures and predefined rules, with its capacity ceiling constrained by the underlying large models and centralized control mechanisms. In contrast, AI-ModelNet seeks to establish an open, standardized, and distributed architecture for model interconnection, where models are abstracted into standardized capability nodes that can be dynamically discovered, on-demand composed, and flexibly invoked to adapt to complex and evolving task environments. The goal of AI-ModelNet extends beyond task-level collaborative execution to fostering the cross-domain flow and global sharing of model capabilities and knowledge. Thus, existing multi-model collaboration systems can be viewed as early implementations and exploratory precursors to AI-ModelNet. It provide crucial technical insights into how models may interact, collaborate, and achieve autonomy, serving as a vital bridge connecting current practices with the future ecosystem of AI-ModelNet.

## 2 AI-Model Network

This section presents the conceptual framework and hierarchical architecture of, along with a demonstration of its application scenarios.

### 2.1 the conceptual framework of AI-ModelNet

AI-ModelNet is a novel paradigm that establishes standardized communication pathways among models, enabling their interconnection, capability sharing, and collaborative reasoning. While it operates on existing Internet infrastructure, there are fundamental distinctions from the conventional Internet in terms of connectivity targets and organizational objectives. Specifically, whereas the traditional internet is built around computing nodes and data resources and depends on universal communication protocols for information transmission, AI-ModelNet takes AI models as its foundational units. It establishes information pathways among heterogeneous models through collaborative protocols. Hence, the internet underpins the communication infrastructure required for model-to-model connectivity. The existing rich heterogeneous models—such as tiny models, giant models, lightweight models, free models, general-purpose models, domain-specific models, and private models—provide the foundational conditions for AI-ModelNet. It features the ability to optimize output results and eliminate individual biases. It offers advantages such as stable service delivery and distributed model updates. Its potential includes emergent intelligence and the decoupling of intelligence from compute power and data. Therefore, while the development of models gives rise to AI-modelNet, it will in turn amplify the value of models.

The schematic diagram of AI-ModelNet is illustrated in Figure 2. Firstly, AI-ModelNet is architected around independently deployed models distributed across multiple edge devices and servers. Secondly, protocols such as HTTP and MCP facilitate interaction between models, establishing dynamic links that collectively form an adaptive model interaction network. AI-ModelNet supports the dynamic entry and exit of models, achieving plug-and-play reasoning. Finally, based on task requirements and real-time network conditions, the system identifies models with compatible capabilities, schedules their execution according to predefined collaboration strategies, and thereby dynamically assembles an optimal logical model pathway. The traditional internet connects computational nodes, while AI-ModelNet connects models and shares knowledge. AI-ModelNet possesses the following characteristics.

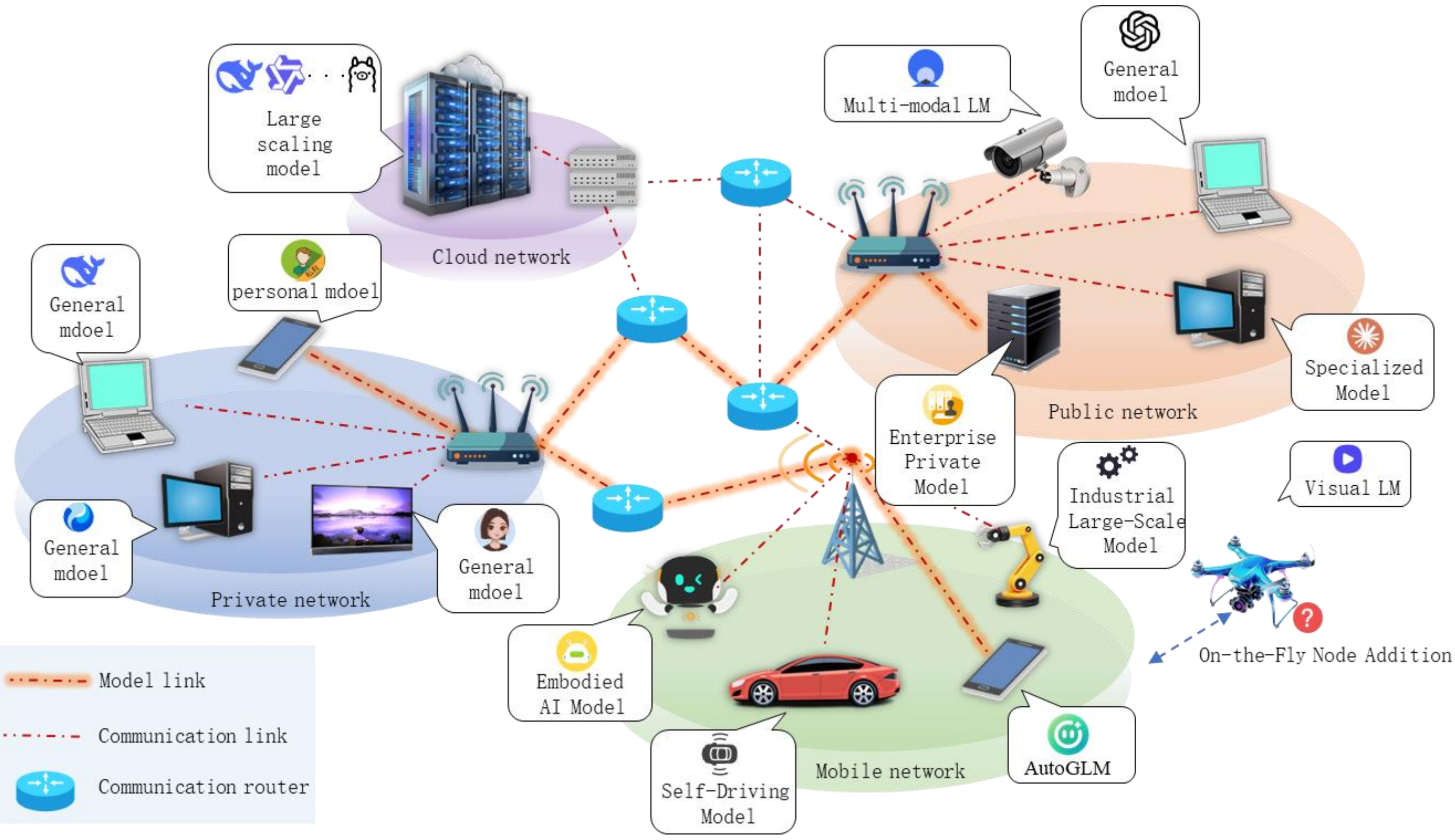

Fig. 2 Schematic diagram of AI-ModelNet

1) Interoperability

Drawing inspiration from the distributed architecture of the Internet, AI-ModelNet incorporates a design philosophy of decentralization and autonomous networking, emphasizing self-governed communication among models. As a result, models within AI-ModelNet are logically interconnected. In this system, models act as network nodes, transmiting messages and sharing resources through a unified model interconnection protocol. Based on task requirements, models dynamically discover other nodes, forming a flexible and scalable collaborative network. The system supports self-organization and topological reconfiguration of nodes, allowing the network to automatically adapt its structure when models go online, offline, or experience performance changes, the collaborative network can automatically adjust its structure to maintain system stability and operational efficiency.

Moreover, the distributed and self-organizing networking mechanism endows AI-ModelNet with enhanced robustness and flexibility. Within this architecture, no fixed master-slave relationships exist among model nodes—any node can assume roles such as task coordination, result fusion, or knowledge sharing when necessary. This dynamic, multi-to-multi, loosely coupled collaborative model breaks free from the hierarchical constraints of traditional model systems, laying the groundwork for achieving inter-domain and cross-platform multi-model coordination.

2) Diversity

By integrating model nodes with diverse capabilities, AI-ModelNet enables collaborative processing of inter-domain tasks, overcoming the knowledge limitations of single models. Different models can undertake specific subtasks based on their expertise, thereby enhancing overall system efficiency and result quality. Such diversified model collaboration grants the system stronger adaptability and flexibility when handling complex and uncertain tasks.

Unlike traditional multi-model systems that only support high-level agent collaboration, AI-ModelNet facilitates cross-granularity cooperation—from macro task-level to micro span-level or token-level cooperation—meeting diverse application requirements for flexibility and precision in collaboration. This multi-model collaboration not only expands task-processing methods but also promotes the interlocking and sharing of knowledge across domains, enabling the system to tackle more complex and varied application scenarios.

3) Dynamic

AI-ModelNet establishes an open, collaborative, and dynamically evolving intelligent ecosystem. Model nodes can join or exit the network at any time, supporting distributed updates and capability evolution, thereby adapting to the rapid iteration demands of models in real-world scenarios.

Meanwhile, the capability requirements of downstream tasks present a dynamically evolving spectrum. Interaction pathways between models are no longer pre-defined, but rather generated and adjusted in real-time along with task flows. Consequently, the overall topology of AI-ModelNet manifests as a continuously reconfiguring dynamic graph network, rather than a fixed, unchanging collaborative architecture.

To summarize, AI-ModelNet enables model nodes to autonomously establish connections and collaboratively accomplish complex tasks through distributed deployment and free networking mechanisms. This architecture enhances system stability, robustness, and scalability while significantly reducing the engineering complexity of model updates and scheduling. Model nodes can directly access capability services provided by other nodes without redeploying all models, thereby realizing dynamic discovery, invocation, and task allocation of model capabilities. By supporting multi-type and multi-granularity collaboration modes, AI-ModelNet provides systemic support for building a flexible, efficient, and sustainably evolving model ecosystem.

## 2.2 Hierarchical Stucture of AI-ModelNet

AI-ModelNet is built upon internet communication protocols and cloud-edge-end computing infrastructures. Various models are deployed across cloud platforms, edge devices, and end devices according to their scale and resource requirements. Their collaboration and interaction are accessible through HTTP-based information exchange and service invocation. As illustrated in Figure 3, AI-ModelNet comprises four layers: the resource layer, the service layer, the interaction layer, and the application layer.

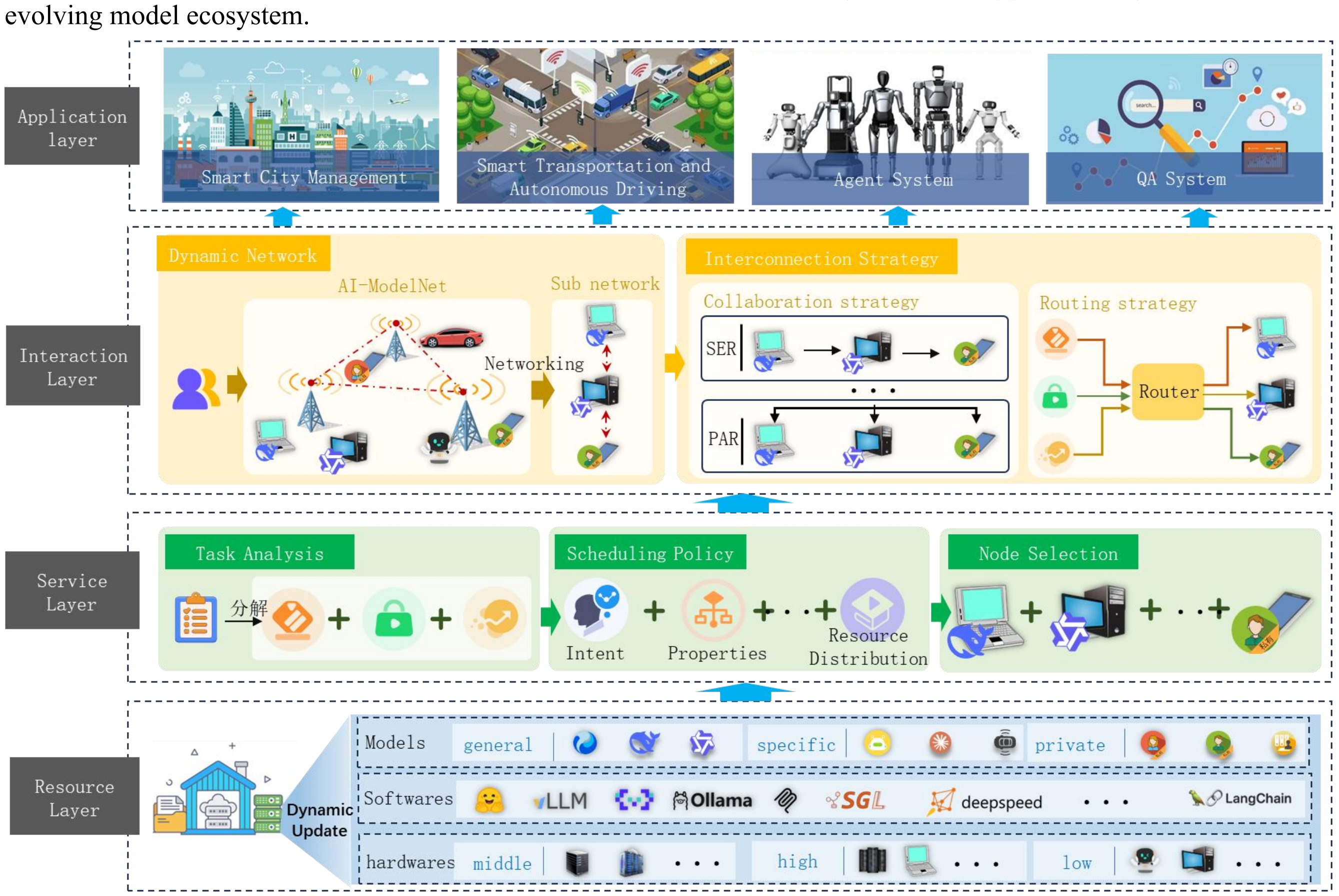

Fig. 3 Hierarchical structure of AI-ModelNet

1) Resource layer

The resource layer serves as the core infrastructure of AI-ModelNet system, responsible for the unified management of various foundational resources. It provides essential support to upper-layer services, ensuring the system operates efficiently and sustainably. Additionally, the resource layer must exhibit high flexibility to accommodate the dynamic entry and exit of diverse models within the system. During model updates or when handling complex tasks, the system should be capable of rapidly adjusting resource allocations. Through effective resource management mechanisms, the resource layer can significantly reduce operational costs while enhancing the user experience.

In AI-ModelNet, the resource layer transcends the traditional scope of servers and network devices typical in cloud computing, evolving into an integrated resource ecosystem that encompasses hardware assets, software environments, and model resources. Hardware resources span heterogeneous computing resources, including high-performance GPU/TPU clusters in the cloud, edge-side AI acceleration nodes, and end-user devices. Software resources are built upon containerization and microservices architectures, integrating middleware such as large model inference engines and service runtime environments to provide standardized, portable, and highly available runtime support for model services. Model resources include cloud-based general-purpose large models as well as various models deployed at the edge or on devices, enabling hierarchical provisioning and on-demand invocation. Grounded in open interfaces, the resource layer allows third-party model providers, computing power operators, and application developers to freely access and share resources, gradually constructing the new collaborative infrastructure of AI-ModelNet.

2) Service layer

The service layer acts as the core hub for intelligent task execution within AI-ModelNet, taking on the dual responsibilities of task orchestration and resource scheduling. Its key functions mainly include decomposing complex tasks into independent subtasks, dynamically updating system resource status, enabling intelligent dynamic scheduling, and balancing the allocation of model workloads. The specific workflow is as follows:

First, the intelligent module within the service layer conducts semantic analysis on tasks submitted by the interaction layer. Through the task analysis module embedded within the service layer, complex requests undergo multi-level decomposition based on task intent and data characteristics. This module automatically breaks down intricate requests into a set of atomic, schedulable subtasks, each tagged with multi-dimensional problem labels. Simultaneously, this layer continuously obtains real-time operational statuses from a vast number of model nodes in the resource layer, including remaining computing capacity, network latency of edge devices and cloud services, and combines this information with model versions and inference delays stored in the global model data center to construct a dynamic, fine-grained resource capability map. Subsequently, the system employs a multi-dimensional label matching mechanism to intelligently align each subtask's requirements with the capability labels of resource nodes, generating a candidate resource set for joint optimization. Finally, it monitors latency fluctuations and anomalies during execution, automatically initiating hot-switching and rescheduling mechanisms in response to node failures or performance degradation to ensure sustained service reliability.

3) Interaction layer

The interaction layer, serving as the core decision-making and control component of the system, dynamically constructs communication network topologies tailored to task characteristics based on the matching results of the service layer. Building on this, the interaction layer continuously adapts its behavior through collaboration and routing strategies, enabling efficient cross-node collaborative inference and data exchange. The specific workflow is detailed below:

The interaction layer first dynamically establishes adaptive collaboration pathways in response to user requests, based on the model-task matching results supplied by the service layer. Subsequently, the interaction layer applies collaboration and routing strategies to achieve efficient coordination across heterogeneous nodes. Collaboration strategies define the cooperative workflow among participating nodes, ensuring efficient and coordinated execution of model computation or inference tasks across heterogeneous computing resources. For example, by introducing serial and parallel collaborative inference algorithms, the system can perform weighted fusion, voting decisions, and debates on outputs from multiple models to achieve unified consensus on inference results. Moreover, routing strategies dynamically determine optimal message transmission paths by analyzing real-time network topology and communication loads, while incorporating multidimensional metrics—such as latency minimization, link stability, and model adaptability—into the route selection process. These strategies aim to enhance data transmission efficiency and overall response performance while minimizing cross-node communication latency and bandwidth overhead. To ensure feasibility, the system adopts HTTP as the baseline transmission mechanism for cross-node communication. Finally, to guarantee compatibility and extensibility, lightweight message encapsulation and state synchronization mechanisms are employed to support high-concurrency, low-latency collaborative model interactions.

4) Application layer

As the tier where ultimate value is realized in AI-ModelNet, the application layer drives the deep integration and efficient collaboration of heterogeneous, cross-platform, and multimodal intelligent models. It aims to build a large-model collaborative infrastructure with self-organizing capability, flexible scalability, and high robustness. Its core value lies in breaking down the barriers of isolated “model silos” that lead to fragmented applications, thereby realigning with the highly complex, dynamically evolving, and multi-source heterogeneous systemic demands of the real world—such as those in scientific exploration, smart manufacturing, and other representative scenarios.

In scientific exploration, AI-ModelNet enables joint solutions to complex scientific problems by integrating diverse types of models, such as physical modeling, data-driven approaches, and knowledge-based reasoning. Whether for climate change prediction or new drug development, the system can automatically schedule and coordinate models across all stages, accomplishing high-dimensional cognitive tasks that are difficult to achieve through traditional methods. In industrial manufacturing, AI-ModelNet supports end-to-end intelligent decision-making in scenarios such as smart manufacturing and product development. It integrates fragmented processes—including defect detection, supply chain optimization, and energy consumption regulation—into a dynamically responsive whole, thereby driving the intelligent transformation of industries.

The continuous expansion of application scenarios will drive AI-ModelNet to deeply penetrate various fields, gradually evolving into a model network where general-purpose and specialized models collaboratively enhance each other. Increasingly diverse application demands propel the evolution of model collaboration mechanisms, while more powerful collaborative capabilities, in turn, further expand the boundaries of new applications. This dynamic evolutionary mechanism positions it to grow into a core infrastructure supporting breakthroughs in cutting-edge scientific research and the intelligent transformation of industries.

### 2.3 Application Examples of AI-ModelNet

This section provides a detailed example of building an AI-ModelNet system, along with experiments conducted on this system to demonstrate its superior performance.

#### 2.3.1 Prototype System of AI-ModelNet

Figure 4 illustrates a prototype system of the Model Internet. Based on technologies such as llama.cpp and distributed cluster management, this system implements management of devices and models within the cluster, as well as the integration and efficient scheduling of heterogeneous devices and model resources across the cluster. Serving as an engineering validation vehicle for the concept of AI-Modelnet, this prototype system provides an experimental platform and foundational validation for subsequent research on model collaboration mechanisms. The following sections will introduce the design and implementation of this system, focusing on model networking implementation, model management, and access control mechanisms.

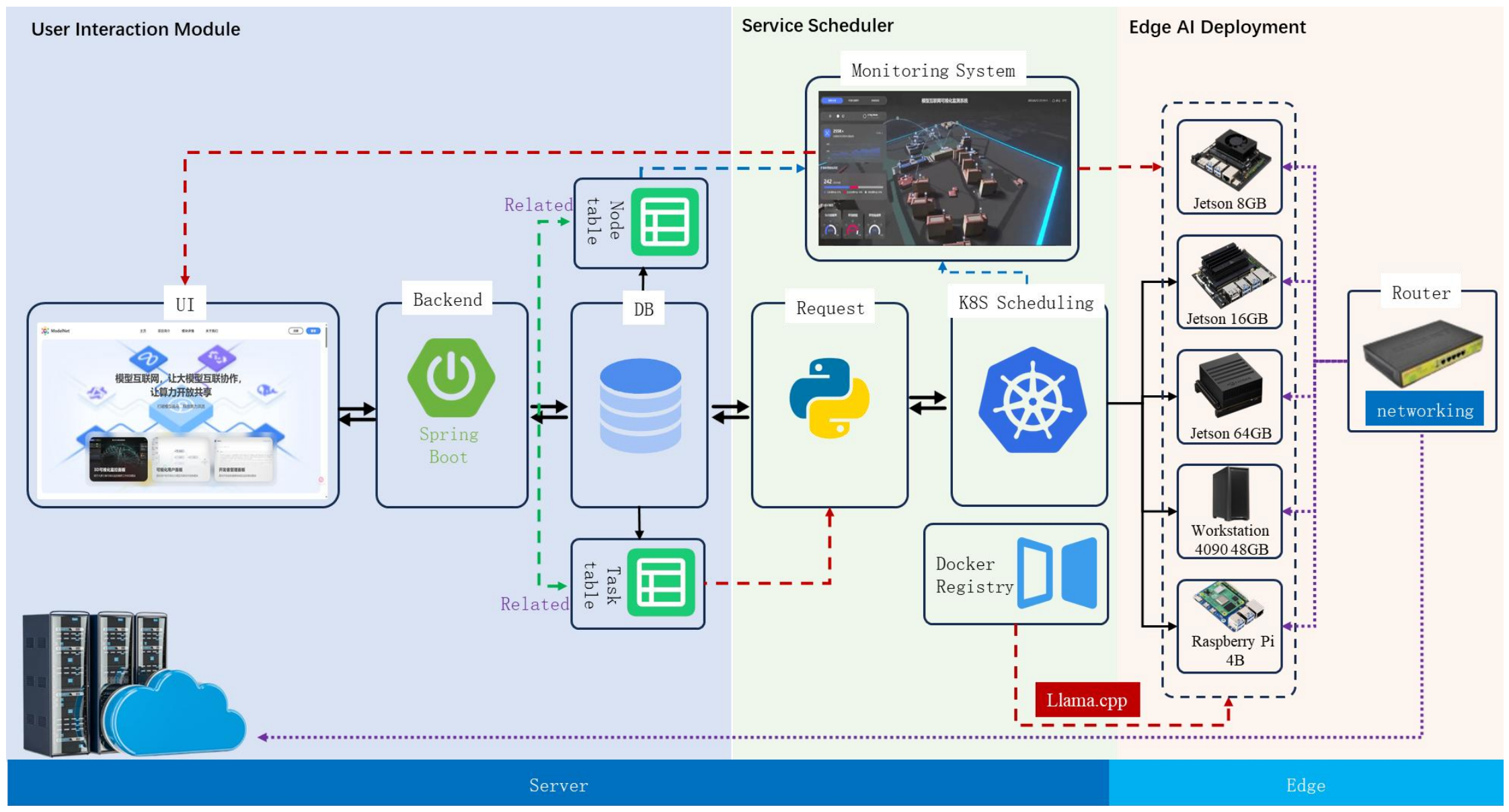

Fig. 4 Prototype system of AI-ModelNet

The system integrates various heterogeneous devices, such as Jetson embedded modules, Raspberry Pi boards, and GPU workstations, as model deployment nodes to simulate diverse computing resources in real-world scenarios. Based on the actual computing power of each node, the system deploys models of corresponding parameter scales, thereby effectively simulating the heterogeneity of model capabilities in a real-world Model Internet.

Furthermore, to facilitate unified management of inference frameworks and model weights across nodes, the platform has developed a comprehensive scheduling solution based on Kubernetes, covering model deployment, model pool construction and inference task scheduling. By encapsulating the llama.cpp framework into standardized Docker images and decoupling model weights from the inference framework, the system achieves separation between resource utilization and model management. This architectural design ensures the overall scalability of the model internet system.

In the prototype system of AI-Modelnet, users first select the desired inference model and collaboration paradigm on the interface, then submit the request information to the scheduling backend. Upon receiving the request, the backend triggers the scheduling process. Next, the scheduler queries the database to obtain relevant model node information and scheduling details, initiates a scheduling request to the scheduling endpoint, and logs the request. Finally, after parsing the response, the scheduling endpoint updates the status of the user's request, writes the inference results to the database, and returns the results to the frontend, thereby completing the service loop.

During the scheduling execution phase, the system first retrieves the deployment status of the target model from the cluster and database. If the model is already deployed on an available node, the system determines whether to accept the request based on the node's real-time resource usage. If the target model is not deployed on any node, the scheduler will select an appropriate idle node for dynamic deployment. To proceed, the node's pre-configured initialization program first extracts the deployment specifications from the scheduling request. It then checks the local cache for the presence of the required model weights as its initial step. If not cached, the program retrieves the model weights and launch parameters from the system's built-in model weight repository, loads the model, completes the deployment, and prepares for subsequent inference.

### 2.3.2 System Verification

To investigate the accuracy, stability, creativity, and ubiquity of AI-ModelNet, this paper selects models independently deployed in the AI-ModelNet system as the research subjects, specifically including: Qwen1.5-7B on Raspberry Pi 8GB, Glm-4-9B on Jetson 16GB, Meta-Llama-3.1-8B on Jetson 64GB, and Mistral-Neom-Instruct-2407 on a workstation with 4090 48GB. To ensure experimental consistency, this paper sets Top-K=3 for answer generation.

The AI-ModelNet system provides standardized interfaces for all connected models, enabling seamless information exchange among them. In routing-based collaboration, the optimal output is selected as the final response based on the model's capability vectors and task attributes. In greedy token-level parallel collaboration, multiple models generate token candidate sets in parallel, and tokens are selected using a greedy strategy that picks the highest-probability token from the candidate pool. Besides, the system integrates the DuetNet [23] inference architecture, which employs a joint truncation strategy to filter token candidate sets from different models and subsequently fuses them to generate optimized results.

1) Accuracy Analysis of AI-ModelNet

As shown in Figure 5, this paper selects objective questions related to photosynthesis. Under the condition of fixed prompts (only requiring the output of answers), the independent responses of the models are compared with the results optimized by the AI-ModelNet greedy token-level strategy to evaluate the improvement in question-answering performance. The correct answer for this question is C. According to the experimental results, it is evident that individual models exhibit shortcomings in answer accuracy and conciseness. For instance, Qwen deployed on small-scale devices produced erroneous conclusions during the analysis process, which compromised the reliability of its responses. Meanwhile, Glm and Mistral-Nemo, while capable of providing detailed analysis, included excessive information in their answers, deviating from the core requirements of the question. Obviously, the output generated by AI-ModelNet under the greedy token-level parallel collaboration strategy showed significant improvements in both accuracy and conciseness, effectively meeting the demands for standard answers. This example demonstrates that model interconnection can integrate the strengths of various models, and through cross-validation and optimized selection of contributions from each model, reduce potential errors and redundancies introduced by individual models.

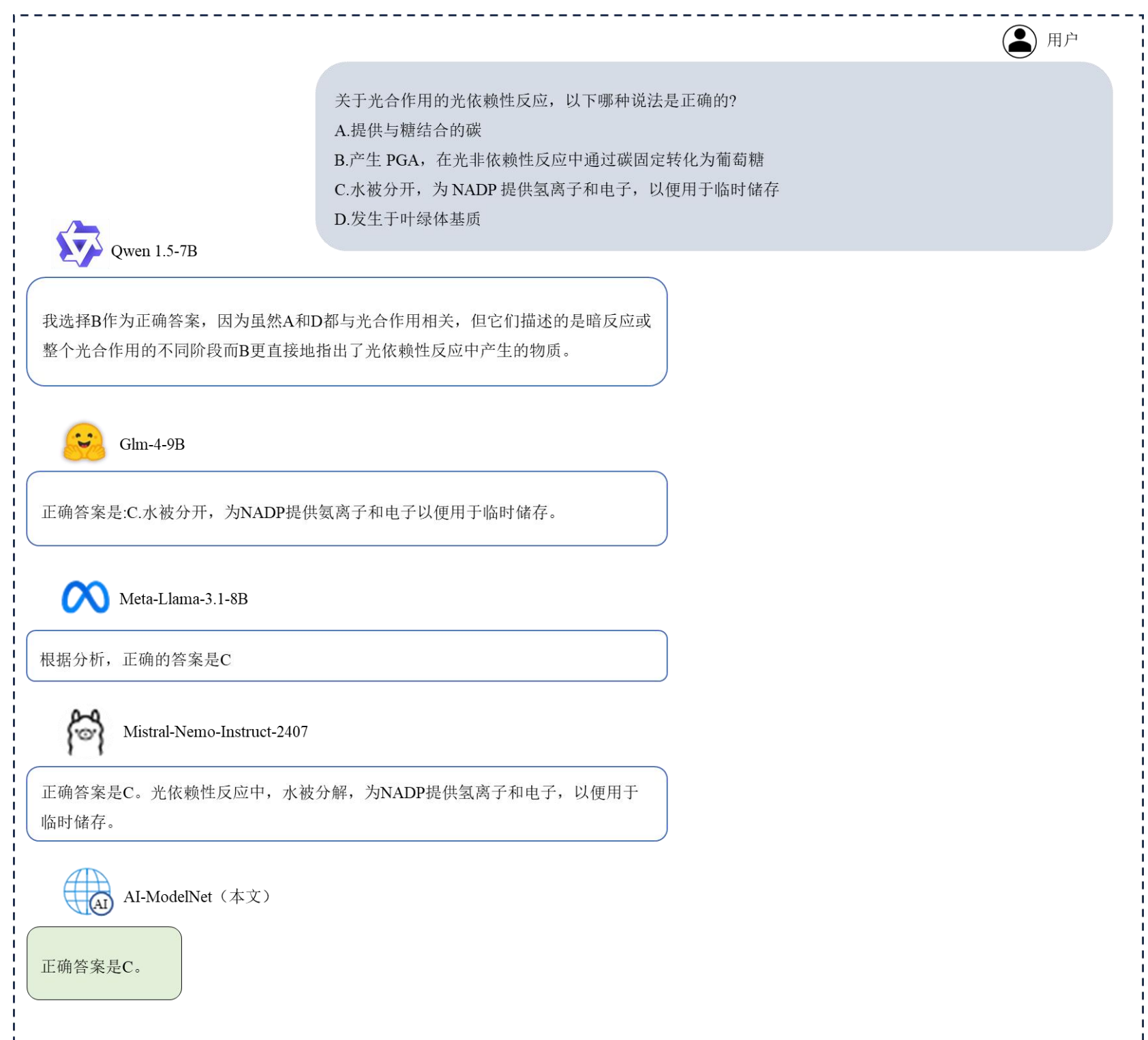

Fig. 5 Comparison of results between single-model and AI-ModelNet

2) Stability Analysis of the Model Internet

The AI-ModelNet system maintains service stability and continuity even during model updates or network disruptions. In the validation experiments, this paper selects models such as Yi-6B, Internlm2-7B, Llama3-8B, Mistral-7B, and Qwen-14B, with unified online rates set at 1, 0.9, 0.8, 0.7, 0.6, and 0.5. The test set is selected from the MMLU dataset. By comparing the answer accuracy of individual models with that of AI-ModelNet under the greedy token-level parallel collaboration strategy, the impact of the interconnection and collaboration mechanism on answer quality and system stability at different online rates is evaluated.

Figure 6 shows that the online rate of individual models exhibits a clear positive correlation with answer quality, meaning that as the online rate gradually increases, the probability of providing high-quality services also rises. The primary reason for this phenomenon is that when individual models fail to provide services normally, it negatively impacts the stability and accuracy of answer quality. However, in AI-ModelNet, the online rate has almost no effect on answer quality. Even if some models disconnect or become unavailable due to network failures, the collaboration mechanism can still allocate and supplement tasks through other available models, thereby providing continuous and stable services. Furthermore, the experimental results indicate that model collaboration not only effectively enhances the system's fault tolerance and robustness but also demonstrates significant advantages in overall answer quality. This fully demonstrates the reliability and superiority of AI-ModelNet in scenarios with low online rates and complex conditions.

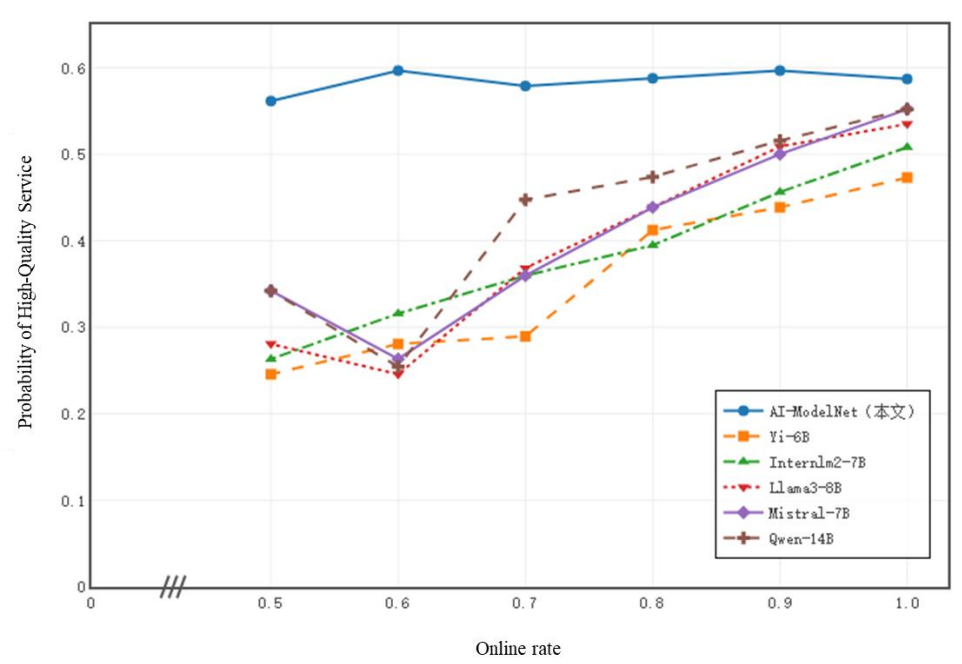

Fig. 6 Stability verification of AI-ModelNet

3) Creativity Analysis of AI-ModelNet

As discussed earlier, individual models exhibit limitations in knowledge performance across different domains due to variations in their training datasets, making it challenging to deliver high-quality responses in certain areas. Additionally, AI models inherently tend to be overconfident—even when lacking relevant knowledge, they are prone to generating seemingly plausible but factually inaccurate content, thereby compromising the authenticity and reliability of their output.

The experimental results are shown in Figure 7. In this example, the correct answer is C. Although each model independently arrived at an incorrect answer, they successfully produced the correct result through collaborative deliberation. This highlights how model interconnection enhances response accuracy, thereby unlocking a higher level of systemic creativity. By leveraging their unique knowledge strengths, the models complemented each other in collaboration, collectively completing the reasoning process. This process not only enhanced the collective capability but also fostered the deep integration and subsequent innovation of knowledge. In tackling complex or cross-domain challenges, the collaborative model mechanism leverages complementary strengths to explore richer solution pathways and expand the possibility space for innovation, ultimately yielding more insightful and creative outcomes. Such multi-model synergy not only improves the system's adaptability to complex tasks but also promotes the diversification and innovation of output content, expanding the potential applications of AI models in knowledge creation and solving real-world problems.

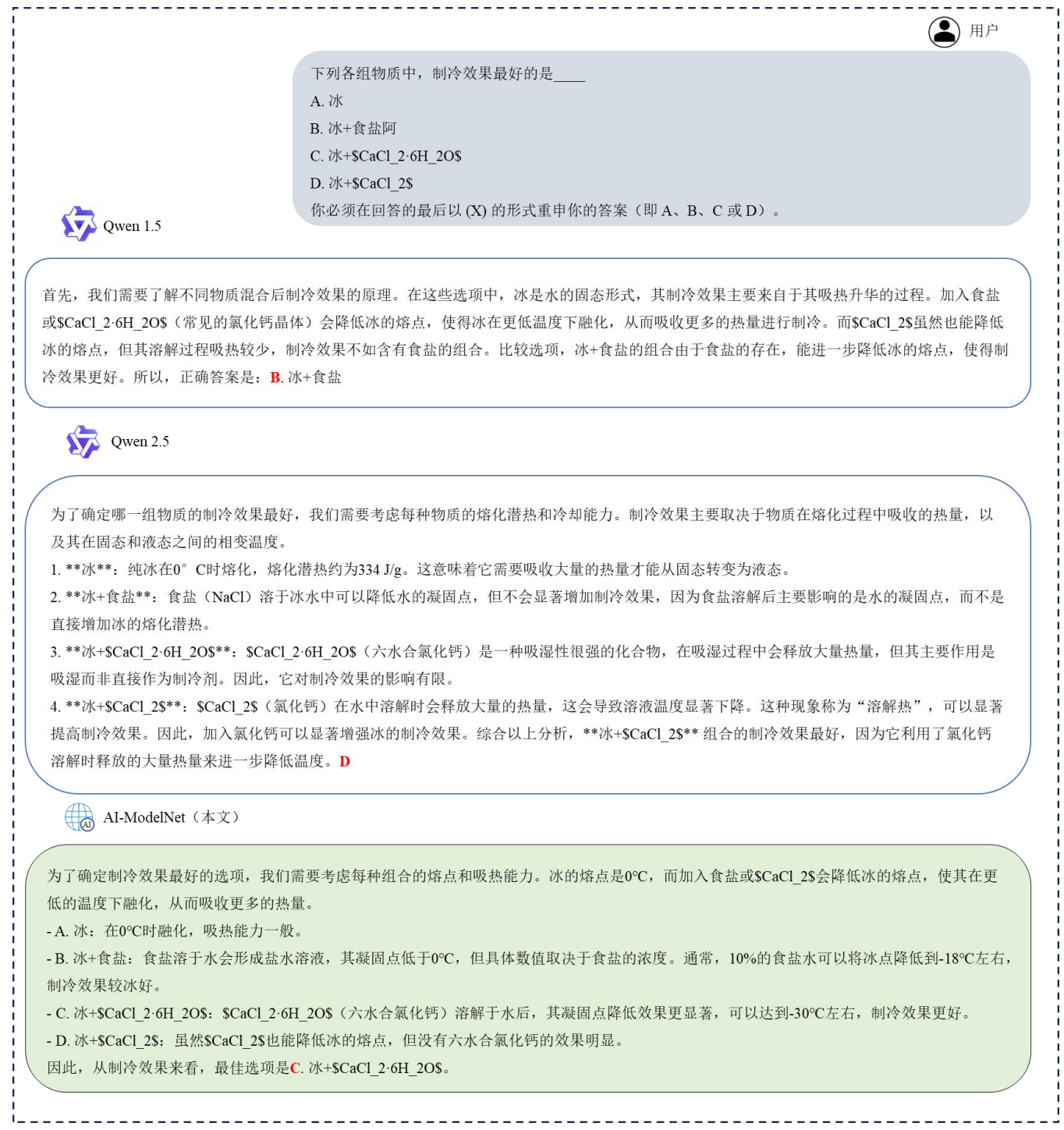

Fig. 7 The creativity of AI-ModelNet

4) Generalizability Analysis of AI-ModelNet

To further explore the generalization potential of the model interconnection network, this study conducted experiments based on a single model (Qwen2.5-7B-Instruct), comparing its reasoning performance before and after integration with the AI-ModelNet. Detailed data are presented in Table 1. After connecting the model to the interconnection network, this paper examines two collaboration mechanisms: token-level collaboration via DuetNet and a typical routing-based collaboration strategy, aiming to systematically evaluate the impact of different collaboration methods on model performance. The datasets used in the experiments are as follows:

- SimpleMath[35]. A dataset for training and evaluating models' basic mathematical problem-solving abilities.
- C-Eval[36]. Used to assess the knowledge and logical reasoning capabilities of large models, comprising multiple-choice questions across 52 disciplines.
- BoolQ[37]. An English sentence-level true/false dataset containing 12,000 entries, designed to investigate models' multi-sentence reasoning ability.
- MMLU[38]. A comprehensive dataset for evaluating models' general knowledge and problem-solving skills, covering multiple-choice questions across 57 disciplines and varying levels of expertise.

Table 1 Accuracy Comparison Between a Single Model and AI-ModelNet %

| Dataset | Before connection | After Connection | |
|---|---|---|---|
| | Qwen 2.5 | Router | DuetNet |
| SimpleMath | 72 | 96 | 92 |
| C-Eval | 70 | 70 | 78 |
| BoolQ | 82 | 84 | 84 |
| MMLU | 86 | 86 | 88 |

The results in Table 1 demonstrate that the accuracy of the interconnected models surpasses that of the standalone model. When employing the routing-based collaboration strategy, the interconnected system achieved a 24% improvement in accuracy on the SimpleMath task compared to the single model. This gain primarily stems from the strategy's ability to prioritize and invoke the most task-specialized model within the model interconnection network. Furthermore, the DuetNet collaboration strategy exhibited broad task adaptability after interconnection. Compared to the standalone model, this strategy increased accuracy by 20%, 8%, and 2% on the SimpleMath, C-Eval, and MMLU datasets, respectively. These improvements arise from the models' capacity to converge on reasoning consensus, thereby reducing error accumulation during inference and enhancing overall reasoning accuracy.

The above result indicates that integrating a model into AI-ModelNet can transcend the inherent limitations of a single model, effectively addressing gaps in task coverage and generalization performance. On the one hand, collaborative interconnection mechanisms can substantially enhance model performance across diverse tasks; on the other hand, the implementation of such collaboration strategies helps reduce users' uncertainty regarding model output performance, thereby improving system reliability and user satisfaction.

The experimental results regarding the accuracy, stability, creativity, and generalization of AI-ModelNet demonstrate that collaboration among models promotes complementarity and enhancement of their knowledge, while offering significant advantages in improving the overall quality of solving complex problems. By integrating the capabilities of multiple models, this collaborative approach enables dynamic resource scheduling and intelligent task allocation, thereby delivering more robust support and stable model services across diverse scenarios. Furthermore, multi-model collaboration effectively mitigates risks associated with single points of failure. Through complementarity and coordinated optimization among models, the system's overall robustness, response speed, and service quality are enhanced, ensuring efficient operation even in complex and changing demand environments to meet users' long-term and stable requirements. Additionally, AI-ModelNet exhibits excellent scalability, allowing iterative upgrades with technological advancements and the introduction of new models, further strengthening collaborative capabilities and providing sustainable and reliable solutions for a wide range of application scenarios.

## 3 Future Research

1) Heterogeneity analysis of models

Various heterogeneous models constitute the AI-ModelNet. Differences among these models primarily stem from factors such as training data distribution, model architecture, and parameter scale. Firstly, models demonstrate significant disparities in their performance across different task types and specialized domains. Secondly, models tend to form internal clusters. Finally, notable differences can be observed in efficiency and resource attributes such as inference latency, inference cost, and context length. Defining the dissimilarities between models is crucial for matching models to tasks.

Existing research categorizes models based on their capabilities using open-source datasets across various domains. For example, Song et al. [39] defined capabilities across 25 dimensions and assigned manual scores to models based on their performance in different tasks. Although capability-based model categorization has been widely adopted and validated in existing studies, defining model heterogeneity still faces the following challenges: Firstly, there is no consensus in the industry on how to establish a unified set of capability dimensions, making it difficult to develop an effective model evaluation framework. Additionally, some evaluation methods still rely on manual intervention, which inevitably introduces subjective biases. Lastly, the coverage of existing open-source datasets is limited, making it challenging to comprehensively encompass all specialized domains. Therefore, defining model heterogeneity is crucial for task matching, and the lack of consensus on heterogeneity has become a significant bottleneck hindering precise model interaction.

2) Model Collaboration Methods

AI-ModelNet aggregates a large number of heterogeneous models and faces a wide variety of task types. Therefore, collaboration among models must be dynamically constructed and adapt in real time according to task requirements.

Existing research predominantly employs pre-defined, fixed models and static interaction patterns, focusing mainly on output-level integration strategies. As a result, it struggles to balance system efficiency in dynamic task scenarios. For instance, TETRIS [40] relies on expert models for continuous guidance, leading to long-term occupation of critical model resources and thus reducing the operational efficiency of the overall system. Most current collaboration methods adopt static and uniform weight allocation strategies, failing to adaptively adjust based on task contexts and the dynamic utility of model outputs. Consequently, existing approaches are difficult to directly apply to the dynamically changing AI-ModelNey.

The dynamism of models, the diversity of tasks, and the uncertainty of network states collectively form a complex dynamic environment that directly impacts collaboration effectiveness. Therefore, adaptive collaboration among models is a key factor in enhancing the performance of the model internet.

3) Dynamic Model Management

Dynamism is a defining feature of the AI-ModelNet, characterized by the autonomous ability of model nodes to join and leave the network. This "plug-and-play" feature offers notable advantages: the system can swiftly schedule other available models to maintain service when certain models become unavailable due to failures, maintenance, network disruptions, or upgrades. Distributed updates and replacements of models can also be efficiently implemented within this environment.

However, such a high degree of dynamism renders traditional static configuration and centralized management paradigms entirely ineffective. The core challenge lies in the inability to predefine the states of all participating nodes, necessitating the handling of an open environment where topology and node capabilities evolve in real-time. Existing research, such as the solution proposed by A2A [41], manages models through a tagging system by having agents submit an "Agent Card" when joining the network. While this method facilitates model discovery and screening, it fundamentally relies on a "self-reporting" mechanism. In an open, dynamic environment that may involve competition or malicious intent, this approach reveals two critical flaws: first, fraudulent "Agent Cards" can pollute the entire model ecosystem, thereby degrading the overall service quality of the system; second, when certain models become unavailable due to network congestion, hardware failures, or involuntary downtime, the system's inability to promptly detect and update their availability status can lead to severe resource wastage.

Therefore, efficient and adaptive dynamic management serves as the operational cornerstone of the Model Internet. It is not only a prerequisite for enabling the "free access and plug-and-play" functionality of models but also the fundamental guarantee for ensuring overall service quality and system reliability.

4) Privacy and Security Issues

AI-ModelNet enables seamless model interaction and collaboration; however, security vulnerabilities in individual model nodes may propagate system-wide through these collaborative links. Compromised model nodes can serve as entry points for attackers within the system, leveraging trust relationships to disseminate malicious commands or false information to other agents—a phenomenon referred to as the "contagion effect" [42]. For example, in sequential collaboration, malicious prompts embedded within instructions can induce models to execute unauthorized commands, potentially leading to the comprehensive compromise of the entire system [43]. Furthermore, existing research has revealed that large models can extract sensitive information from user prompts [44], and the open nature of AI-ModelNet may amplify the risks of model privacy leakage.

To build a secure and trustworthy model-interconnected ecosystem, the following challenges must be addressed: first, how to identify malicious models to curb the transmission of harmful information at its source; second, how to prevent the leakage of users' sensitive information during interactions. In response to these challenges, future research must focus on advancing two key frontiers: on one hand, developing dynamic defense systems capable of real-time detection and containment of malicious models; on the other hand, overcoming critical technologies to protect data privacy in complex interactions and eliminate the risk of sensitive information leakage. This is the essential path toward achieving a truly secure and trustworthy AI-ModelNet.

5) Heterogeneous Model Interconnection and Interoperability

AI-ModelNet aims to construct a ubiquitously interconnected network of models, enabling the efficient transmission and sharing among heterogeneous models. However, unlike the communication forms of nodes in the traditional internet, AI-ModelNet lacks pre-established communication links; its "connections" are transient, dynamic, and soft-linked. Furthermore, the organizational structure among models also influences the selection of model routing paths.

Existing multi-agent communication protocols offer valuable insights for enabling communication among models, yet significant challenges remain. For example, the Agent Communication Protocol (ACP) [45] is an open standard based on RESTful APIs designed to support agent communication. However, in complex scenarios involving large-scale deployment, this standard still faces two core challenges: first, the lack of a unified framework for dynamic negotiation, information sharing, and model coordination; second, the absence of a cross-domain, mutually recognized trust system to ensure secure collaboration.

AI-ModelNet aims to build an intelligent interconnection system across different entities and platforms, enabling free discovery, autonomous collaboration, and information exchange among models. Therefore, research on model interconnection must not only explore collaborative methods between models but also establish universal, secure, trustworthy, and efficient communication mechanisms at the protocol level.

## 4 Conclusion

AI-ModelNet represents a new paradigm for networked artificial intelligence models. This paper proposes the concept, framework, and hierarchical architecture of AI-ModelNet, along with the development of a related prototype system. Experimental results demonstrate its strong performance in terms of accuracy, stability, creativity, and generalization capabilities. AI-ModelNet aims to explore new modes of efficient interconnection, intelligent interoperability, and complementary capabilities among models, with the goal of advancing large-scale model technology toward a networked direction and providing novel insights for research in the field of artificial intelligence. Furthermore, this paper offers a preliminary outlook on promising research directions within AI-ModelNet, serving as a reference for deepening and expanding future studies in this area.

# 模型互联网: 概念、现状和未来

李哲涛　曾曦玉　王建辉　肖　勇　刘忠仁　吴俊儒　赖俊杰　黄纪俊　龙赛琴
(暨南大学信息科学技术学院　广州　510632)
(liztchina@hotmail.com)

# AI-Model Network: Concept, Current State and Future

Li Zhetao, Zeng Xiyu, Wang Jianhui, Xiao Yong, Liu Zhongren, Wu Junru, Lai Junjie, Huang Jijun, and Long Saiqin
(*College of Information Science and Technology*, *Jinan University*, *Guangzhou* 510632)

**Abstract**　While the primary function of computers lies in computation and processing, the core value of the Internet is rooted in sharing and collaboration. Computers create the Internet, and the Internet empowers the value of computers. The rapid development of the Internet, cloud computing, and big data is pushing artificial intelligence into the era of large models (LMs). However, the practical application of LMs is currently hindered by high training costs and deployment complexities, driving a shift toward lightweight, private, and domain-specific models. With the rapid proliferation and wide distribution of heterogeneous models, enabling effective interaction and collaboration among them has emerged as a critical bottleneck that urgently needs to be addressed in LM development. Drawing inspiration from the development of the Internet, we propose the concept, vision, and system architecture of world wide AI-model network (AI-ModelNet). It is a novel paradigm that achieves interconnection, capability sharing, and collaborative reasoning by establishing pathways between models. We first briefly review the current state of single-model and multi-model research. Subsequently, the systemic vision and hierarchical architecture of AI-ModelNet are articulated, followed by validation of the framework's feasibility through a prototype system and diverse application cases. Finally, key directions for future research are discussed preliminarily.

**摘　要**　计算机的主要功能在于计算与处理,互联网的价值在于共享与协作。计算机催生互联网,互联网放大计算机的价值。互联网、云计算与大数据的快速发展推动了人工智能迈入大模型时代。然而,大模型在实际应用中面临训练成本高昂、泛在部署困难等挑战,促使大模型朝轻量化、私有化及垂直化方向发展。随着异构模型的急速增长和广泛分布,模型间的交互协同成为大模型发展亟须突破的瓶颈问题。受计算机互联网的发展模式启发,提出模型互联网(AI-ModelNet)的概念、构想及系统架构,它是一种通过构建模型间通路,实现互联互通、能力共享与协同推理等的新范式。首先,简要介绍了单一模型与多模型场景的发展现状;然后,系统性地阐述了模型互联网的总体构想与层次结构,并结合原型系统与相关应用实例验证了其可行性与实用性;最后,初步展望了未来研究的主要方向。

**收稿日期**: 2025-03-24; **修回日期**: 2026-01-19
**基金项目**: 国家自然科学基金国际合作重点项目(W2411053); 国家自然科学基金联合基金重点项目(U23B2027)
This work was supported by the Key Program of Projects of International Cooperation and Exchanges NSFC (W2411053) and the Key Project of the Joint Funds of the National Natural Science Foundation of China (U23B2027).

计算机的主要功能在于计算与处理，互联网的价值在于共享与协作。在计算机的发展历程中，随着技术的进步与广泛普及，其研究重心逐步由单机算力增强转向网络通信机制。在 1974 年，TCP/IP 协议的问世为网络报文传输提供了标准，奠定了现代互联网的基础。1989 年，HTTP 协议的提出推动了万维网[1]的兴起，彻底改变了人类社会的信息交流方式，数字空间、信息空间和网络空间等成为人类社会信息交流的重要媒介。因此，计算机催生互联网，互联网则持续拓展计算的应用边界。计算机和互联网相辅相成，互联放大了计算价值。

在人工智能（artificial intelligence, AI）的发展中也同样经历了从简单任务到复杂任务的发展路径。人工智能模型的发展大致经历了 3 个阶段：神经网络阶段、深度学习阶段以及大规模预训练模型阶段。在神经网络阶段，模型结构相对简单，主要用于解决基本的分类和回归问题，为深度学习奠定了根基；在深度学习阶段，网络结构逐步加深，模型能够处理更复杂的数据特征和任务需求，并被广泛应用于图像识别、语音识别和自然语言处理（NLP）等领域；Transformer 架构[2]的提出标志着人工智能步入了大规模预训练模型阶段，在显著提升并行计算效率的同时实现性能突破。如图 1 所示，计算机的发展实现了从简单计算到复杂计算，而互联网实现了从孤立计算到全球互联的范式跃迁。受互联网发展的启发，模型互联有望实现模型价值在系统层面的放大。

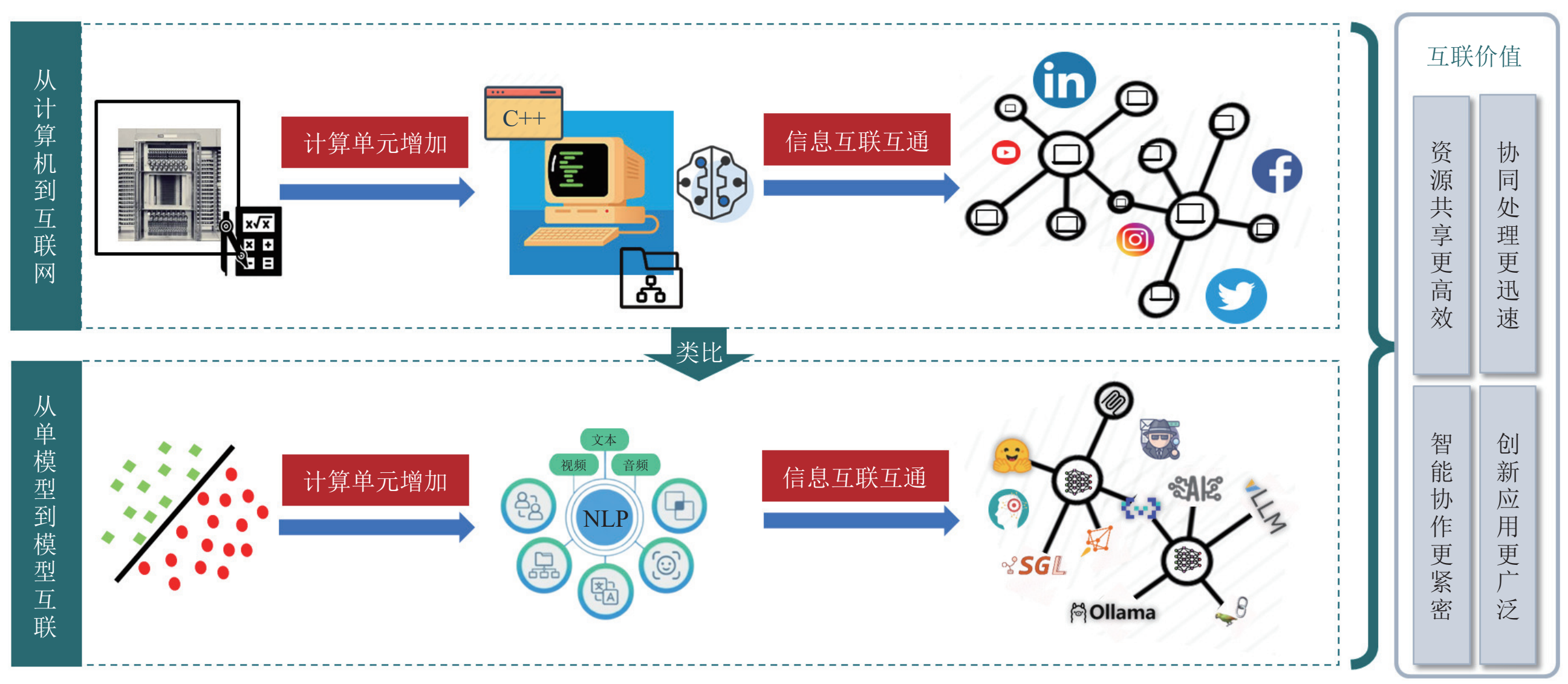

Fig. 1 From computer network to AI-ModelNet

图 1 从计算机网络到 AI-ModelNet

当前的人工智能处于参数爆炸时代。然而，现有人工智能模型面临训练硬件成本、数据成本和人力成本高昂，部署存储资源消耗大且参数规模红利微弱的困境[3]。因此，轻量化、私有化与垂直化部署成为主流趋势。当前模型之间割裂缺少互联，“模型孤岛”现象逐渐显现。在实际应用中，单一模型面临多重挑战。比如模型往往会无意识地继承训练数据的偏见[4]，从而影响其决策的客观性；在知识盲区或不确定场景下，模型可能因自身偏好和幻觉输出不实信息[5]。通用模型在领域数据上进行针对性训练后，通用能力往往会有所下降[6]。正如计算机互联实现资源共享与协同处理以提升整体系统价值，模型互联互通也有望成为迈向通用人工智能（artificial general intelligence, AGI）的重要路径。

在复杂系统中，高阶能力的形成往往并非源于单一主体能力的无限增强，而是多个有限个体在相互联系与协同作用中涌现的结果[5-9]。这一规律在自然系统、社会系统以及工程系统中均具有普遍性：个体能力存在不可逾越的边界，而个体之间的联系则为信息互补、认知纠偏与能力放大提供了结构性条件。从这一视角上看，系统的智能水平更多取决于其组织方式与交互机制，而非单个组件的最优性能。该系统性规律在经验层面早已被广泛认知，常被概括为“三个臭皮匠，顶个诸葛亮”，即集体智慧在适当的组织与协作条件下，往往能够超越个体的能力上限。其本质在于多个体交互带来的协同增益。不同个体在知识背景、推理路径与决策偏好上的差异，使系统能够在交互过程中不断修正局部偏差，弥补认

知盲区，并在整体层面形成更具鲁棒性的判断结果[10]。模型协作的价值并不是抹平每个“盲人”个体的感知局限，而是多元互证中构建坚实共识，通过动态整合局部感知不断趋近“大象”整体真相。因此，个体之间的联系并非附属关系，而是高阶能力产生的必要条件。

将这一系统性视角引入人工智能领域，可以发现模型之间的互联并非偶然选择，而是一种内在必然。单一模型无论在规模或性能上如何提升，其内部表征仍受限于结构容量与数据质量，而模型之间的联系则为跨模型知识互补与推理协同提供了可能性。通过构建模型之间的稳定联系，使多个模型在协作中形成群体智能，能够在系统层面突破单一模型能力的固有边界。这也是“普遍联系”“量变质变”哲学原理的具体体现，即多模型随着协作深度的增加，最终有望推动智能系统在认知层次、任务泛化与自主协同等方面实现“质”的跃迁。统计显示，目前全球已有超过 1 300 个人工智能大模型，涵盖主要知识领域[11]。 如何有效组织与协调现有模型，以互联推动模型网络化，已成为提升智能系统整体效率的重要方向。

如图 1 所示，若将模型视为计算机中封装的智能资源，则模型间的交互过程可类比为网络节点间的信息传输。因此，本文提出“模型互联网（AI-ModelNet）概念”。模型互联网是一种通过构建标准化模型间信息通路，实现模型的动态发现、按需组合与协同推理等的新范式。模型互联网强调模型节点之间的互联互通与网络级协作，将模型抽象为可持续发现与组合的能力节点，以支撑开放、异构且可演化的智能系统模型互联互通实现了人工智能与大算力、大数据的解耦，有效降低智能技术对算力与数据资源的强依赖性。因此，模型互联网有望成为未来人工智能发展的催化剂。

## 1 研究现状

本节依次探讨现有单一模型以及多模型协作领域的研究进展。通过系统性的分析阐明模型互联网与现有研究路径的根本差异。

### 1.1 单一模型

随着大规模预训练技术的快速发展，单一大模型在自然语言处理（NLP）、信息检索等任务中展现出显著优势。伴随 Transformer 架构的演化，众多性能的生成式大模型相继涌现，形成“百模竞发”的发展格局。代表性模型如 GPT 系列[12]、LLaMA 系列[13]、DeepSeek 系列[14]、Qwen 系列[15] 与 Kimi 系列[16] 模型，其应用拓展至语言生成、知识问答、推理与工具调用等多种场景，并在自动驾驶[17]、智能助手[18]、文学创作[19] 等领域得到探索与实践。

单一大模型在实际应用中仍面临诸多挑战。首先，模型的知识与价值边界受训练数据影响。其次，模型可能存在“幻觉”现象，即逻辑连贯但事实错误的信息输出[20]。此外，大模型推理依赖巨大算力、存储与能耗资源[21]，限制了其在资源受限场景中的部署。最后，特定任务的专有训练与微调的模型，往往会以削弱其通用性能为代价。

面对多样化任务需求、复杂环境适配与系统长期演进等挑战时，单一大模型架构在泛化性、高效性与灵活性等方面的瓶颈日益凸显。这促使研究者开始探索具备协同能力的多模型架构，通过任务分解、模型互补与灵活路由等机制，实现能力模块化、推理透明化与模型更新自由化，从而为人工智能研究提供新的发展方向。

### 1.2 多模型协作

近年来，多模型协作逐渐成为构建灵活、高效智能系统的重要路径。多模型协作的核心思想旨在通过模型间能力互补、任务分解与结构化路由，实现复杂任务的分治推理，并增强系统稳健性。现有研究主要集中于 3 类策略[22]：模型融合、输出集成与动态路由。

模型融合通过整合多个预训练模型的参数以构建统一模型，追求结构上的协同与性能压缩。Ties-Merging 方法[23] 则引入轻量策略以缓解参数符号冲突和幅度失衡问题。Twin-Merging 方法通过将专家模型的知识分解为共享知识与独占知识，然后根据输入动态融合两类知识，有效解决了知识干扰和测试数据异构导致的性能下降问题[24]。

输出集成聚焦于在保留模型独立性的前提下，通过输出融合实现更稳健的推理。DeePEn 方法[25] 提出双向词汇映射机制以打通异构模型的语义表示空间。王建辉等人[26] 提出一种 token 级并联协作方法，通过通信机制连接多个模型节点，实现跨设备的高效交互推理。吴俊儒等人[27] 设计的互评串联协作机制实现模型间自治答案优化。此外，多源知识融合框架[28] 也展现出多模型协作在任务分工与知识整合方面的优势。Wu 等人[29] 构建了面向协作的多模型系统，通过角色分配实现智能体间的能力互补，从而提升系统整体性能。

动态路由机制通过在推理过程中动态选择最合适的模型执行子任务，实现性能与成本的平衡，进一步增强了多模型系统的灵活性。Ong 等人[30] 提出的 RouteLLM 基于人类偏好数据引导路由器学习模型选择决策。BEST-Route 方法[31] 结合模型与采样策略进行双重选择。GraphRouter 方法[32] 采用图结构建模模型与查询间的语义关联。RCR-Router 方法[33] 引入角色感知记忆机制实现上下文信息的精准传递。TO-Router 方法[34] 则在工业场景部署中验证了基于查询特征与专家能力映射的高效选择机制。

尽管上述方法在多模型架构设计方面取得了积极进展，但当前研究仍面临三大挑战：一是协作任务设计缺乏通用性，仍依赖特定场景配置；二是模型间交互机制固化，难以适配动态多任务的实际环境；三是系统高度依赖集中算力资源，导致部署成本高和扩展性不足。虽然现有多模型协作研究已奠定了良好基础，但其在开放、异构与分布式环境中的适用性仍有待提升。这也进一步凸显了“模型间互联”作为通用基础设施的重要价值，即：通过建立松耦合、可通信及可协作的全球模型互联网络体系，突破当前协作范式的边界，推动智能系统向高灵活性、高扩展性与长期自治方向演进。

综上所述，多模型协作系统与模型互联网均以实现模型间的协作与能力互补为目标，但二者在范式预设上存在本质差异。前者多依赖静态架构与预定义规则，其能力上限受制于底层大模型与集中式控制机制；而模型互联网则致力于构建一个开放式、标准化及分布式的模型互联架构，其中模型被抽象为可被动态发现、按需组合与灵活调用的标准化能力节点，以适应复杂多变的任务环境。模型互联网的目标不仅在于任务层面的协同执行，更在于促进模型能力与知识的跨域流动与全局共享。因此，现有的多模型协作系统可被视为一种实现模型互联网的早期实践与前期探索。它为模型如何交互、协同与自治提供了重要技术思路，是连接当前实践与未来模型互联网生态的重要桥梁。

## 2 模型互联网

本节提出了模型互联网的构想及其层次结构，并展示了模型互联网的应用实例。

### 2.1 模型互联网的构想

模型互联网是一种通过构建模型间标准化通路，实现模型互联互通、能力共享与协同推理的新范式。模型互联网依托现有互联网基础设施运行，但其连接对象和组织目标与互联网不同：互联网以计算节点与数据资源为核心，通过通用通信协议实现信息传输；而模型互联网以人工智能模型为节点单元，基于协作规则在异构模型之间构建信息通路。因此，互联网为模型互联提供通信基础支撑。现有微小模型、巨大模型、低配模型、免费模型、通用模型、领域模型和私有模型等丰富异构模型为模型互联网提供了基础条件。模型互联网具有优化输出结果、消除个体偏见等特点；提供稳定服务、模型分布式更新等优点；智能涌现、智能与算力/数据解耦等潜能。因此，是模型的发展催生模型互联网，模型互联网也将放大模型的价值。

模型互联网的示意图如图 2 所示。模型互联网是由若干边端设备以及服务器上私有化部署的模型组成。它可应用 HTTP, MCP 等协议，使得模型之间进行交互形成通信链路，由此构建出动态的模型交互网络。模型互联网允许自由地接入或退出模型，实现推理的“即插即用”。它根据任务性质与实时状态，遴选出能力匹配的模型，进而依据预设的协作策略进行调度与执行，从而在逻辑上动态形成一条最优的模型链路。互联网连接的是计算节点；模型互联网连接的是模型，分享的是知识。模型互联网具有以下特征：

1）互通性

借鉴了互联网的分布式网络结构，模型互联网引入了分布式与自由组网的设计理念，强调模型间的自治通信。因此，逻辑上模型互联网中的模型是互联互通的。在该体系中，模型作为网络节点通过统一的模型互联通信协议实现消息传递与共享。模型依据任务需求动态发现其他节点，从而形成灵活可扩展的模型协作网络。系统支持节点的自组织与拓扑重构，当模型上线、离线或性能变化时，协作网络可自动调整结构以保持系统稳定与运行高效。

此外，分布式的自由组网机制赋予了模型互联网更高的鲁棒性与弹性。在此架构下，模型节点之间不存在固定的主从关系，任何节点在必要时承担任务协调、结果融合或知识共享的角色。这种动态多对多、弱耦合的协作模式打破了传统模型系统的层级约束，为实现跨域、跨平台的多模型协同奠定了基础。

2）多样性

模型互联网通过组织具备差异化能力的模型节点，实现跨领域任务的协同处理，而不再依赖单一模型的知识边界。不同模型可依据其专长承担特定子

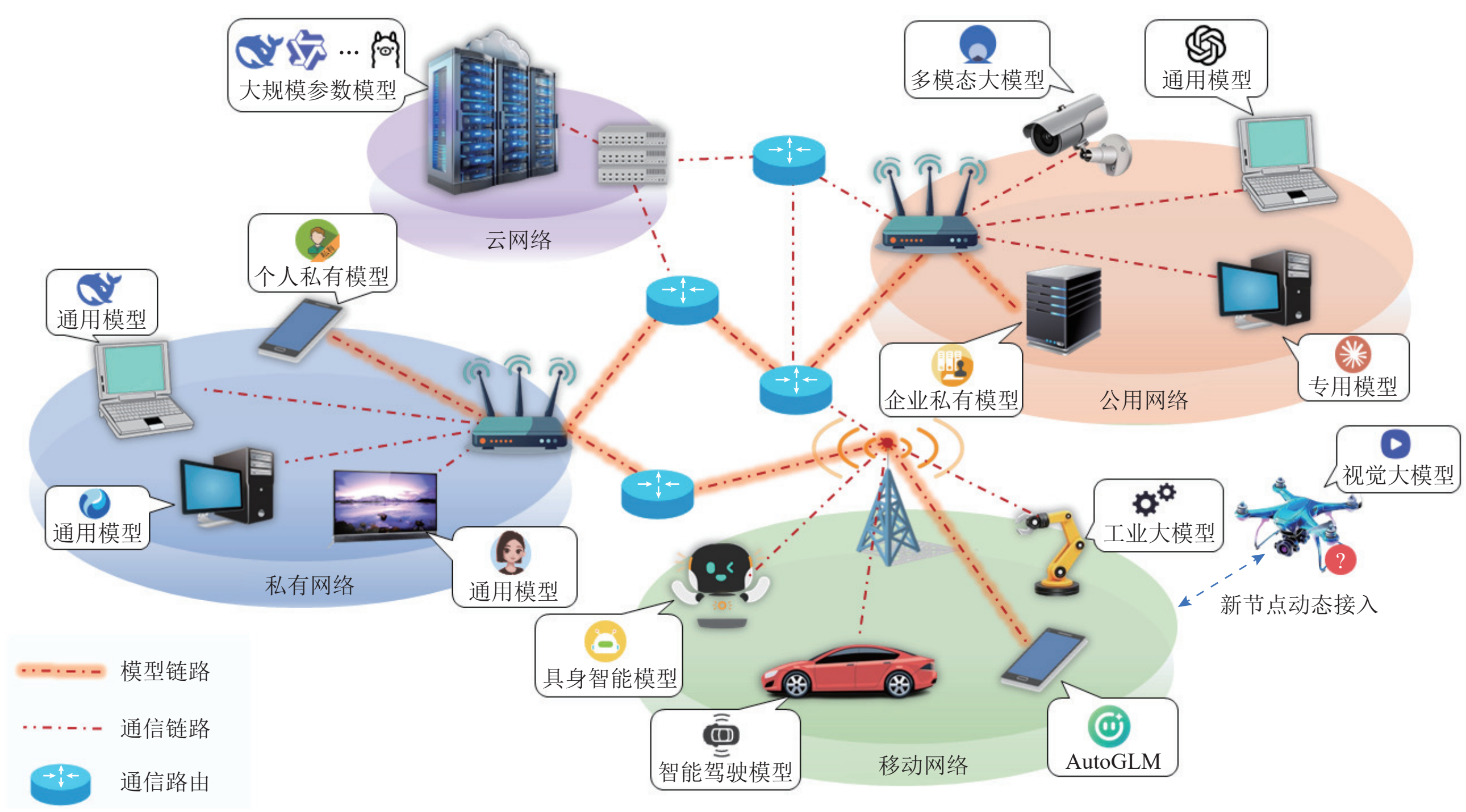

Fig. 2 Schematic diagram of AI-ModelNet
图 2 AI-ModelNet 的示意图

任务,从而在系统层面提升整体效率与结果质量。这种多样化的模型协作使系统在面对复杂与不确定任务时,具备更强的适应性和灵活性。

不同于传统多模型系统仅支持高层 agent 级协作,模型互联网支持从宏观任务级到微观 span 级或 token 级的跨粒度协作,能够满足不同应用场景对协作灵活性和精细度的要求。这种多模型的协作不仅扩展了任务处理方式,还促进了不同领域知识的联动与共享,从而能够应对更复杂和更多样化的应用场景。

3)动态性

模型互联网构建了一个开放、协同及动态演化的智能生态系统。模型节点可随时接入或退出网络,支持分布式更新与能力演进,从而适应模型快速迭代的现实需求。

与此同时,下游任务对模型能力的需求呈现出动态变化的能力谱系。模型之间的交互链路不再是预先定义的,而是随着任务流实时生成与调整。由此,模型互联网的整体拓扑结构表现为一个持续重构的动态图网络,而非固定不变的协作架构。

综上所述,模型互联网通过分布式部署与自由组网机制,使模型节点能够自主建立连接并协同完成复杂任务。该体系在增强系统稳定性、鲁棒性与扩展能力的同时,显著降低了模型更新与调度的工程复杂度。模型节点可直接访问其他节点提供的能力服务,无需重复部署全部模型,从而实现模型能力的动态发现、调用与任务分配。通过支持多类型、多粒度的协作模式,模型互联网为构建灵活、高效且可持续演化的模型生态系统提供了系统性支撑。

### 2.2 模型互联网的层次结构

模型互联网以云边端物理设备与互联网通信协议为基础。各类模型根据其规模与资源需求分别部署于云端、边缘设备及端侧设备中,模型间的协同与交互则体现为基于 HTTP 协议的信息交换与服务调用。如图 3 所示,模型互联网由资源层、服务层、交互层以及应用层组成。

1)资源层

资源层作为模型互联网系统的核心基础设施,负责对各类基础资源进行统一管理。其为交互层提供了必要的支持,确保系统高效且可持续的运行。此外,资源层需要具备高度灵活性,以适配各类模型自由接入和退出系统的需求。在模型更新或处理复杂任务时,系统应能够迅速调整资源配置。同时,通过有效的资源管理机制,资源层可以有效降低运营成本,提升用户体验。

在模型互联网中,资源层不再局限于传统云计算应用中的服务器与网络设备,而是拓展为涵盖硬件资源、软件资源与模型资源的综合性资源集合。

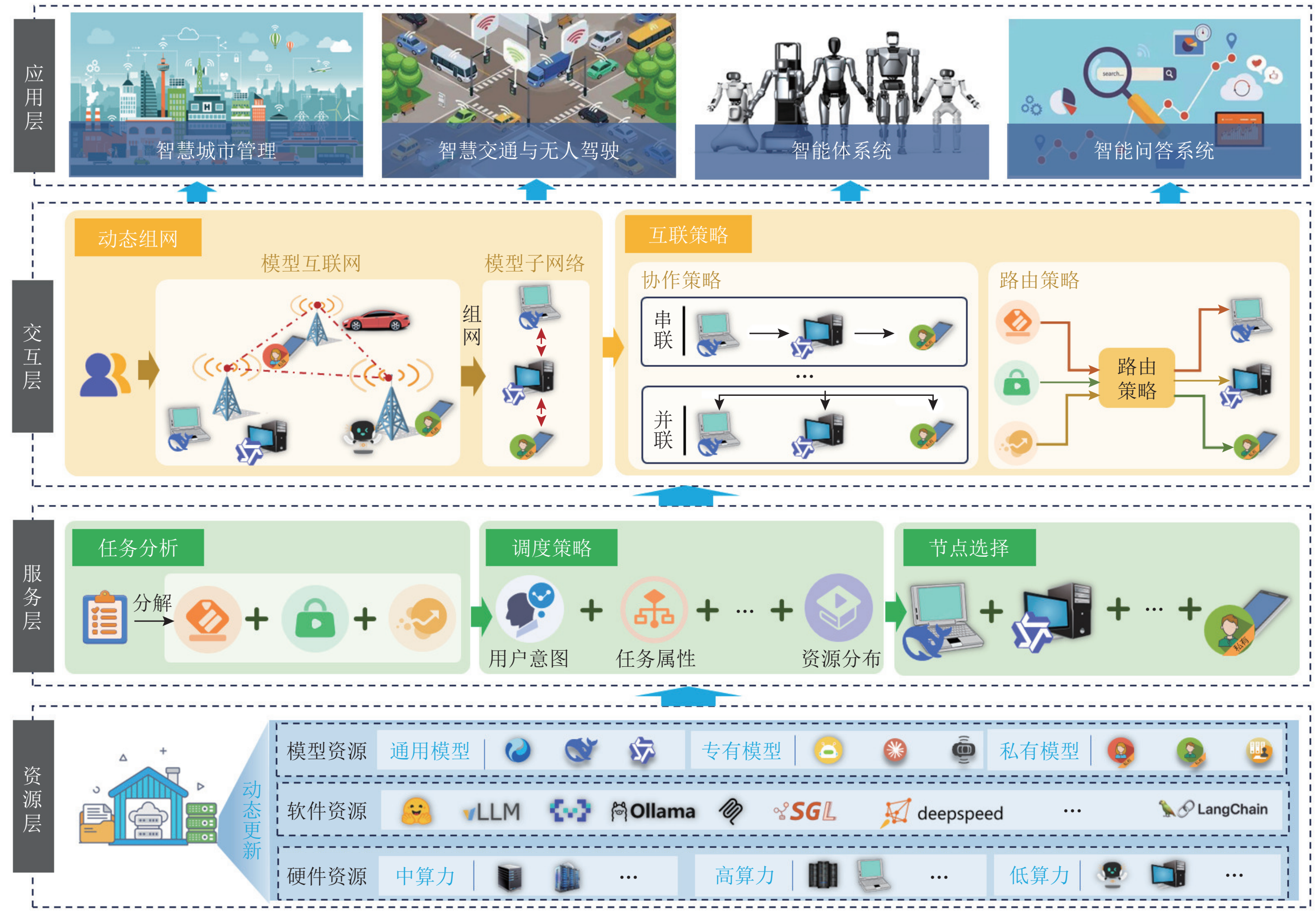

Fig. 3 Hierarchical structure of AI-ModelNet

图 3 AI-ModelNet 的层次结构

其中,硬件资源则覆盖云端高性能 GPU/TPU 集群,边缘侧 AI 加速节点以及终端设备等异构计算资源。软件资源则构建于容器化与微服务架构之上,整合了大模型推理引擎、服务运行环境等中间件,为模型服务提供标准化、可移植、高可用的运行时环境。模型资源包括云端通用大模型以及各种边端部署的模型,用于实现模型能力的分层供给与按需调用,该资源通过模型池进行管理。资源层以开放式接口为基石,支持第三方模型提供方、算力运营商与应用开发者自由接入与共享资源,逐步构建起模型互联网的新型协作基础设施。

2)服务层

服务层是模型互联网中实现智能任务执行的核心枢纽,承担着任务编排与资源调度的双重职责。其核心功能主要包括将复杂任务分解为独立的子任务、动态更新系统资源状态、智能动态调度以及模型负载的均衡分配。具体工作流程如下:

首先,服务层的智能模块需要对交互层所提交任务进行语义分析。通过服务层内置的任务分析模块对任务意图、数据特征等进行多层次拆解,将复杂请求自动分解为一组原子化、可调度化的子任务,并为每个子任务打上多维问题标签。同时,服务层持续从资源层获取海量节点的实时运行状态,包括各边缘设备与云服务的算力余量、网络延迟等,并结合全局模型数据中心中存储的模型版本、推理延迟等信息,构建出一张动态、细粒度的资源能力图谱。然后,通过多维度标签匹配机制,系统将每个子任务的需求与资源节点的能力标签进行智能匹配,形成候选资源集合,并进行联合优化匹配。最后,监控执行过程中的延迟波动与异常状态,在节点失效或性能劣化时自动触发热切换与重调度机制,保障服务可靠性。

3)交互层

交互层作为系统决策与控制的核心环节,需要根据服务层输出的匹配结果,动态构建面向任务特性的通信网络拓扑结构。在此基础上,交互层依据协作策略与路由策略做出动态调整,实现跨节点的高效协同推理与数据交互。具体工作流程如下所示:

交互层首先根据服务层输出的模型-任务匹配结果,针对用户发起任务请求动态构建自适应的协作

通路。随后，交互层应用协作策略与路由策略，实现跨异构节点的高效协同。其中，协作策略用于定义各参与节点在任务执行过程中的协作流程，确保模型计算或推理任务在算力异构的节点环境中实现高效协同。例如，通过引入模型串行与并行协作推理算法，可对多模型输出进行加权融合、投票决策和辩论等，实现各类模型推理结果的统一共识。路由策略基于实时网络拓扑状态与通信负载动态规划消息传输路径，综合考虑最短延迟、链路稳定性、模型适配性等多维指标进行路由选择。路由策略以提升数据传输效率与整体响应性能为目标，尽可能减少跨节点通信延迟与带宽开销，系统可采用 HTTP 协议作为跨节点通信的基线传输机制。此外，为确保兼容性与可扩展性，通过轻量级消息封装与状态同步机制以支撑高并发、低延迟的模型协作需求。

4)应用层

应用层是互联网模型的价值实现层，负责将跨类型、跨平台、多模态智能模型的深度融合与高效协同，封装为一个可适配各类任务场景的统一接口。其核心价值在于打破“模型孤岛”导致的单一化应用的壁垒，满足现实世界中高度复杂、动态演化以及多源异构的系统性需求，如科学探索、智能制造等典型场景。

在科学探索中，模型互联网能通过整合物理建模、数据驱动与知识推理等不同类型的模型，实现对复杂科学问题的联合求解。无论是气候变化预测，还是新药研发，系统均能自动调度与协调各环节模型，完成传统方式难以企及的高维认知任务。在产业制造中，模型互联网系统能支撑智能制造、产品研发等场景下的全链路智能决策，将缺陷检测、供应链优化、能耗调控等分散环节整合为动态响应的整体，实现产业智能化升级。

应用场景的持续拓展，将驱动模型互联网向各领域深度渗透，逐步演进为通用模型与专有模型协同增强的模型网络。日益丰富的应用需求推动模型协同机制的进化，而更强大的协同能力又进一步拓展新的应用边界。这一动态演进机制，使其有望成长为支撑前沿科学研究突破、产业智能化升级的核心基础设施。

### 2.3 模型互联网的应用示例

本节详细描述了一个模型互联网系统搭建的实例，并通过在该系统上的实验展现了其良好性能。

#### 2.3.1 模型互联网原型系统

图 4 展示了一个模型互联网原型系统。该系统基于 llama.cpp 和分布式集群管理等技术，实现了对集群内设备和模型的管理，以及对集群内异构设备与模型资源的整合与高效调度。作为模型互联网概念的工程化验证载体，该原型系统为后续模型协作机制的研究提供了实验平台与验证基础。下面将围绕模型互联实现、模型管理及接入控制机制，介绍该系统的设计与实现。

该系统集成了 Jetson 嵌入式模块、树莓派与 GPU 工作站等多种异构设备作为模型部署节点，以模拟真实场景中多样化计算资源。该系统根据各节点的实际算力，为其部署相应参数规模的模型，从而有效模拟了真实模型互联网中模型能力的异构性。

此外，为便于跨节点统一管理推理框架与模型权重，该平台基于 Kubernetes 构建了一套完整的调度方案，涵盖模型的部署、模型池的构建与分发、推理任务调度。通过将 llama.cpp 框架封装为标准化的 Docker 镜像，并实现模型权重与推理框架的分离，系统在资源利用与模型管理之间实现了解耦，从而在架构层面保障了模型互联网系统的整体可扩展性。

在模型互联网的原型系统中，用户首先在界面上选择想要使用的推理模型和协作范式，并将请求信息提交到调度后端，当后端接收到请求后触发调度流程。接着，调度器查询数据库，获取相关模型节点信息与调度信息，向调度端发起调度请求，并记录该请求日志。最后，调度端解析响应后，更新用户请求状态，将推理结果写入数据库，并向前端返回结果，完成服务闭环请求。

在调度执行阶段，系统首先在集群与数据库中检索目标模型的部署状态。若模型已部署于可用节点，则根据该节点的实时资源占用情况决定是否接受请求；若无节点部署该模型，调度器将选择一个合适的空闲节点执行动态部署。随后，当目标节点启动时，其预置的初始化程序从调度请求中提取待部署模型信息，并优先检查本地缓存是否已存在对应权重。若未缓存，程序将从系统内置的模型权重库中拉取模型权重及启动参数，加载模型并完成部署，为后续推理做好准备。

#### 2.3.2 系统验证

为探究模型互联网的精准度、稳定性、创造性以及泛在性等，本文选取独立部署在 AI-ModelNet 系统中的模型作为研究对象，具体包括：树莓派 8 B 上的 Qwen1.5-7B、Jetson 16 GB 上的 Glm-4-9B、Jetson 64 GB 上的 Meta-Llama-3.1-8B、4090 工作站 48 GB 上的 Mistral-Neom-Instruct-2407。为确保实验的一致性，

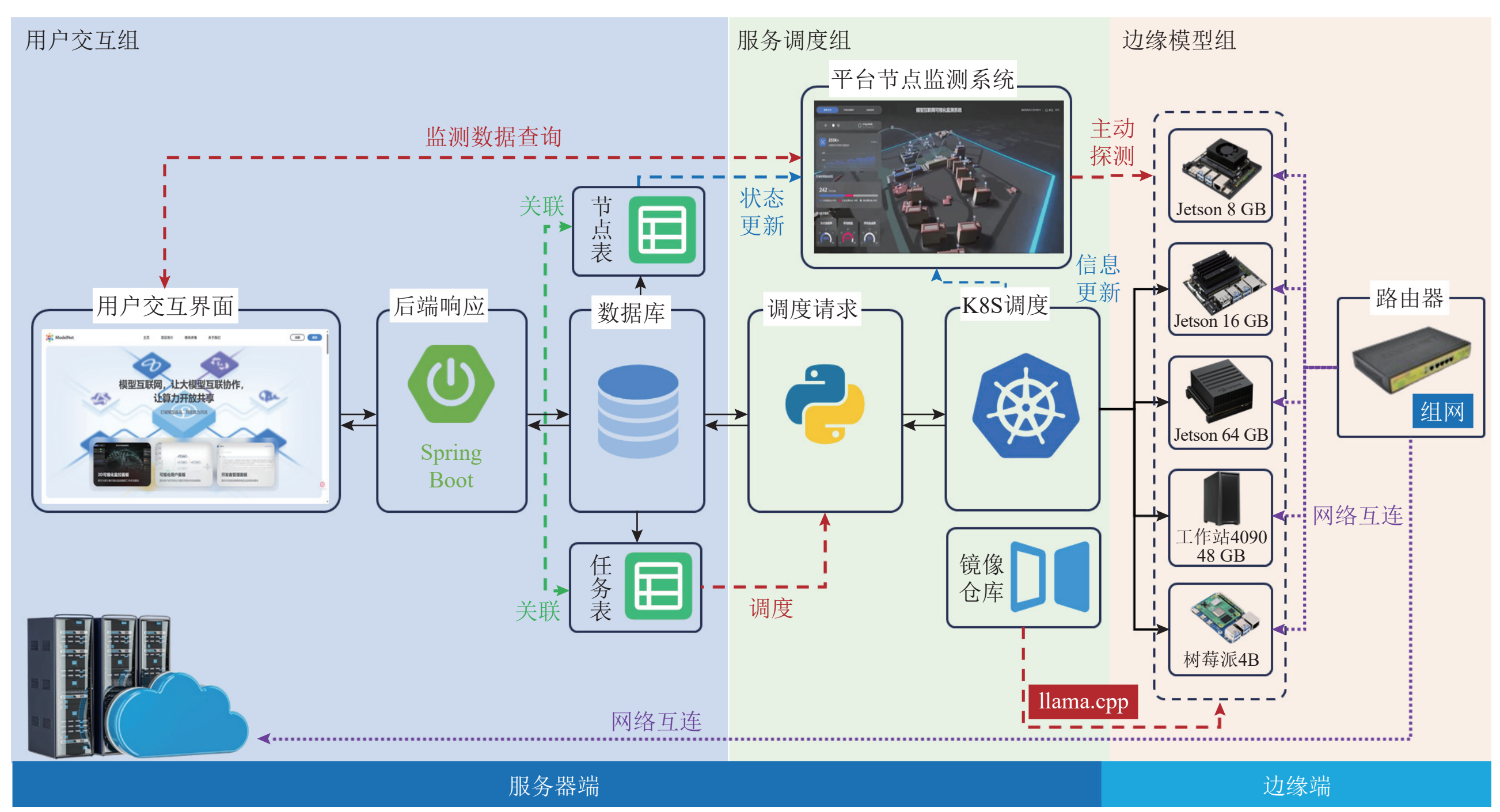

Fig. 4 Prototype system of AI-ModelNet

图 4 AI-ModelNet 的原型系统

本文设置 *Top-K*=3 生成回答。

AI-ModelNet 系统为所有接入模型提供标准化接口,实现了模型间的无缝信息交换。在路由式协作方面,根据模型的能力向量和任务属性选取最优输出作为最终回答。在贪心 token 级并联方式方面,多模型并行输出 token 候选集,根据贪心策略选取候选集中概率最大的 token。另外,系统集成了 DuetNet[23] 推理架构,其通过联合截断策略筛选来自各模型的 token 候选集,进而融合生成优化结果。

1)模型互联网的精准度分析

如图 5 所示,本文选取光合作用相关的客观问题,在固定提示词(仅要求输出答案)的条件下,对比了模型独立作答与经 AI-ModelNet 贪心 token 级策略优化后的结果,以评估问答性能的提升。该问题的正确结果为 C,根据实验结果可知单一模型在答案准确性和简洁性上存在不足。部署在小设备上的 Qwen 在分析过程中产生错误结论,影响了回答的可靠性。然后,Glm 和 Mistral-Nemo 虽然能够提供详细分析,但在回答中包含多余信息,偏离问题的核心需求。显然 AI-ModelNet 在贪心 token 级并联策略上的输出结果在准确性和简洁性上明显提升,能够满足问题对标准答案的要求。从该示例可以发现模型互联能够整合各模型的优点,并通过交叉验证和优化选择模型贡献的部分,减少了单一模型可能带来的差错和冗余。

2)模型互联网的稳定性分析

模型互联网面对模型更替或网络中断等情况,其协同机制依然能够保障服务的稳定性和连续性。在验证性实验中,本文选取了 Yi-6B, Internlm2-7B, Llama3-8B, Mistral-7B, Qwen-14B 等模型,统一设置在线率为 1, 0.9, 0.8, 0.7, 0.6, 0.5, 测试集选择 MMLU 数据集。通过对比单一模型与 AI-ModelNet 在贪心 token 级并联策略的回答准确性,评估互联协作机制在不同在线率下对回答质量和系统稳定性的影响。图 6 表明单一模型的在线率与回答质量呈现明显的正比关系,即当在线率逐步提升时,提供高质量服务的概率也增大。这一现象的主要原因是单一模型无法正常提供服务,影响了回答质量的稳定性和准确性。然而,在模型互联网中,在线率对回答质量几乎没有影响。即使部分模型断开连接或因网络故障无法提供服务,协作机制仍然能够通过其他可用模型进行任务分配和补充,从而提供持续稳定的服务。此外,实验结果进一步表明,模型协作不仅有效提升了系统的容错能力和鲁棒性,还在整体回答质量方面展现出显著的优势。这充分证明了模型互联网在低在线率和复杂场景时的可靠性和优越性。

3)模型互联网的创造性分析

通过前文分析可知,单一模型由于训练数据集的差异,在不同领域的知识表现上存在局限性,难以在某些领域提供高质量的回答。此外,LMs 本身具

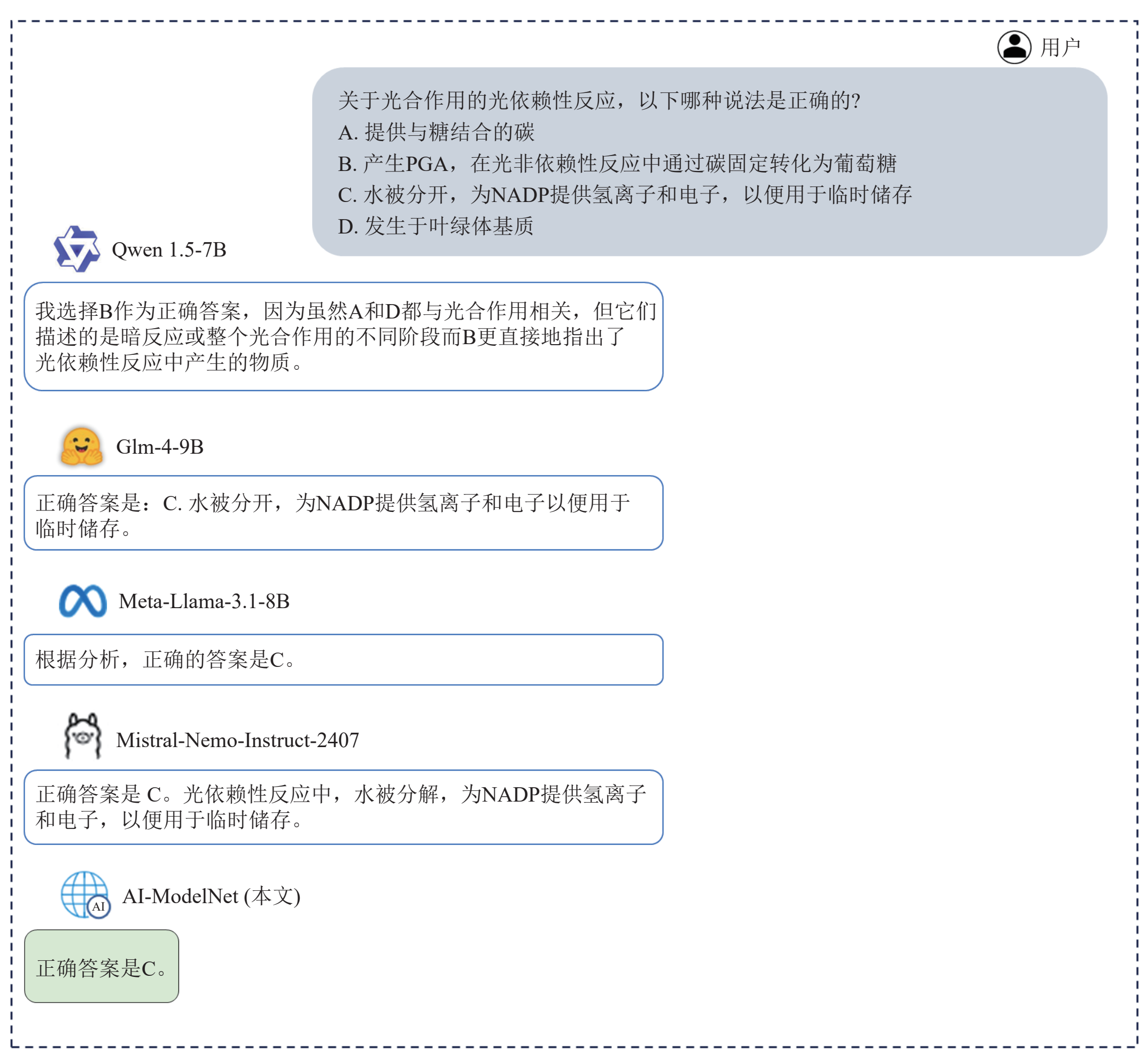

Fig. 5 Results comparison of singlemodel and AI-ModelNet

图 5 单模型和 AI-ModelNet 的结果对比

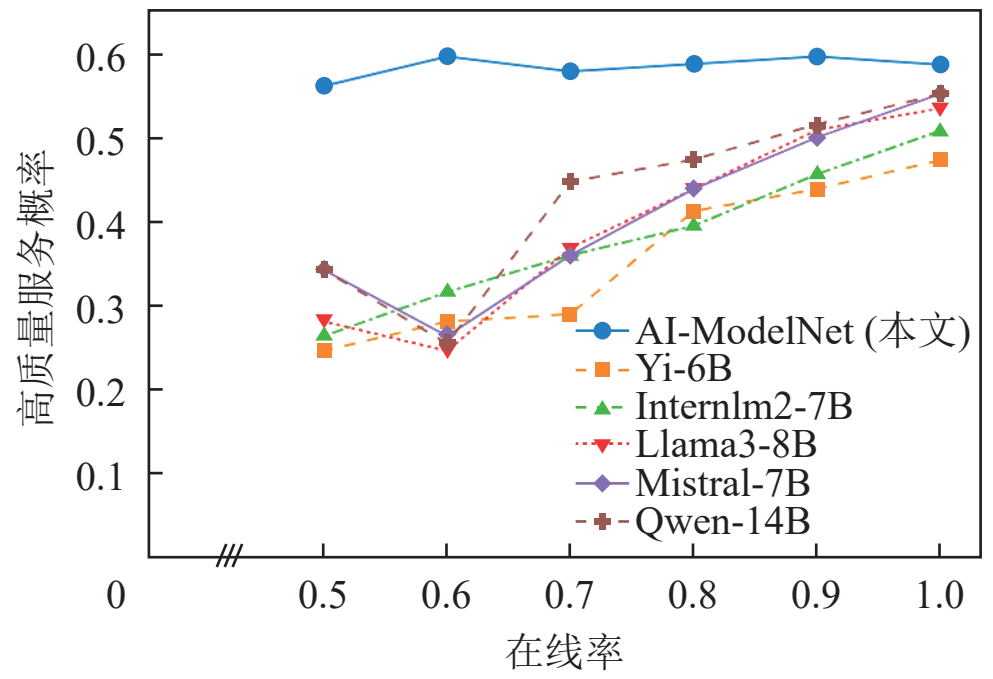

Fig. 6 Stability verification of AI-ModelNet

图 6 AI-ModelNet 的稳定性验证

有过度自信的倾向——即使缺乏相关知识，也倾向于生成看似合理但实际不准确的内容，从而损害输出的真实性和可靠性。为验证模型互联所激发的创造性，本文通过对同系列模型 Qwen1.5-7B，Qwen2.5-7B 在 AI- ModelNet 贪心 token 级并联策略下的协作结果分析模型互联所带来的创造潜力。选择同系列模型旨在控制架构范式与训练数据分布这一核心变量，以探究交互能否催生非预期创新。

实验结果如图 7 所示，在该示例中，正确答案为 C。尽管 2 个模型单独运行时均给出了错误答案，但当它们通过协作互联后，却成功输出了正确结果。这不仅表明模型互联能够提升回答的正确性，也进一步增强了系统的创造性。各模型通过发挥自身独特的知识优势，在协作中相互配合、共同完成推理，不仅优化了整体性能，还促进了知识的深度融合与创新。结果进一步表明，在处理复杂或跨领域问题时，模型协作机制能够有效利用不同模型之间的互补性，探索更加丰富多样的解题路径与创新方案，从而生成更具启发性和创造力的内容。这种多模型协同不仅提高了系统对复杂任务的适应能力，还推动了输出内容在多样化与创新性方面的提升，拓展了人工智能模型在知识创造与解决实际问题中的应用潜力。

4）模型互联网的泛化性分析

为进一步探索模型互联网的泛化潜力，本研究基于单一模型（Qwen2.5-7B-Instruct）开展了实验，比

用户

下列各组物质中，制冷效果最好的是____
A. 冰
B. 冰+食盐
C. 冰+$CaCl_2·6H_2O$
D. 冰+$CaCl_2$
你必须在回答的最后以 (X) 的形式重申你的答案 (即A、B、C或D)。

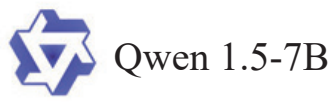
Qwen 1.5-7B

首先，我们需要了解不同物质混合后制冷效果的原理。在这些选项中，冰是水的固态形式，其制冷效果主要来自于其吸热升华的过程。加入食盐或$CaCl_2·6H_2O$ (常见的氯化钙晶体) 会降低冰的熔点，使得冰在更低温度下融化，从而吸收更多的热量进行制冷。而$CaCl_2$虽然也能降低冰的熔点，但其溶解过程吸热较少，制冷效果不如含有食盐的组合。比较选项，冰+食盐的组合由于食盐的存在，能进一步降低冰的熔点，使得制冷效果更好。所以，正确答案是：B. 冰+食盐

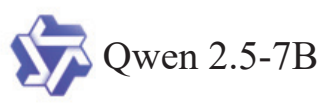
Qwen 2.5-7B

为了确定哪一组物质的制冷效果最好，我们需要考虑每种物质的熔化潜热和冷却能力。制冷效果主要取决于物质在熔化过程中吸收的热量，以及其在固态和液态之间的相变温度。
1. **冰**：纯冰在0 ℃时熔化，熔化潜热约为334 J/g。这意味着它需要吸收大量的热量才能从固态转变为液态。
2. **冰+食盐**：食盐 (NaCl) 溶于冰水中可以降低水的凝固点，但不会显著增加制冷效果，因为食盐溶解后主要影响的是水的凝固点，而不是直接增加冰的熔化潜热。
3. **冰+$CaCl_2·6H_2O$**：$CaCl_2·6H_2O$ (六水合氯化钙) 是一种吸湿性很强的化合物，在吸湿过程中会释放大量热量，但其主要作用是吸湿而非直接作为制冷剂。因此，它对制冷效果的影响有限。
4. **冰+$CaCl_2$**：$CaCl_2$ (氯化钙) 在水中溶解时会释放大量的热量，这会导致溶液温度显著下降。这种现象称为“溶解热”，可以显著提高制冷效果。因此，加入氯化钙可以显著增强冰的制冷效果。综合以上分析，**冰+$CaCl_2$**组合的制冷效果最好，因为它利用了氯化钙溶解时释放的大量热量来进一步降低温度。D

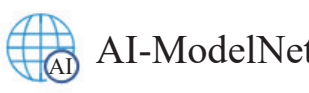
AI-ModelNet (本文)

为了确定制冷效果最好的选项，我们需要考虑每种组合的熔点和吸热能力。冰的熔点是0 ℃，而加入食盐或$CaCl_2$会降低冰的熔点，使其在更低的温度下融化，从而吸收更多的热量。
-A. 冰：在 0 ℃时融化，吸热能力一般。
-B. 冰+食盐：食盐溶于水会形成盐水溶液，其凝固点低于0 ℃，但具体数值取决于食盐的浓度。通常，10%的食盐水可以将冰点降低到−18 ℃左右，制冷效果较好。
-C. 冰+$CaCl_2·6H_2O$：$CaCl_2·6H_2O$ (六水合氯化钙) 溶解于水后，其凝固点降低效果更显著，可以达到−30 ℃左右，制冷效果更好。
-D. 冰+$CaCl_2$：虽然$CaCl_2$也能降低冰的熔点，但没有六水合氯化钙的效果明显。
因此，从制冷效果来看，最佳选项是 C. 冰+$CaCl_2·6H_2O$。

Fig. 7 The creativity of AI-ModelNet

图 7 AI-ModelNet 的创造性

较其在接入 AI-ModelNet 前后的推理性能变化，具体数据如表 1 所示。在模型接入模型互联网后，本文分别考察了 2 种协作机制：token 级协作的 DuetNet 和典型的路由式协作策略，以系统评估不同协作方式对模型性能的影响。

实验采用的数据集如下：

SimpleMath[35] 用于训练和评估模型基础数学应用题能力的数据集。

C-Eval[36] 用于评测大模型的知识和逻辑推理能力，涉及 52 个学科的多选题。

BoolQ[37] 段落级英文判断题，包含 12 000 条数据，用于探究模型的多句推理能力。

MMLU[38] 用于全面评估模型的通用知识和问题的解决能力，涵盖了 57 个学科以及不同专业程度的多选数据集。

由表 1 结果可见，模型互联后的整体准确率优于

Table 1 Accuracy Comparison Between a Single Model and AI-ModelNet

表 1 单一模型与 AI-ModelNet 的准确率对比 %

| 数据集 | 互联前 | 互联后 | |
|---|---|---|---|
| | Qwen 2.5 | 路由式 | DuetNet |
| SimpleMath | 72 | 96 | 92 |
| C-Eval | 70 | 70 | 78 |
| BoolQ | 82 | 84 | 84 |
| MMLU | 86 | 86 | 88 |

单一模型。当模型采用路由式协作策略时，与单一模型相比，该方式在 SimpleMath 任务上的准确率提升了 24 个百分点，这主要得益于其能够优先调用模型互联网中与该任务最契合的模型。其次，DuetNet 协作策略在互联后表现出广泛的任务适应性。相较于未互联的单一模型，该策略在 SimpleMath, C-Eval, MMLU 数据集上的准确率分别提升了 20 个百分点、8 个百分点、2 个百分点。该提升源于模型间通过汇聚推理共识降低推理过程的错误累积，从而提高推理准确率。上述结果表明，接入模型互联网能够突破单一模型的能力边界，有效弥补其在任务覆盖与泛化性能方面的局限。一方面，互联协作机制可显著增强模型在多样化任务上的表现；另一方面，协作策略的实施有助于降低用户对模型输出性能的不确定性，从而提升系统可靠性与用户满意度。

以上对模型互联后的准确性、稳定性、创造性以及泛化性的实验结果表明：模型间的协作可以推动模型知识的互补与增强，并且在提高复杂问题解答的整体质量上具有显著优势。这种互联协作模式通过集成多个模型的能力，实现动态资源调度与智能任务分配，从而在多样化场景下提供更稳定的模型服务支持。同时，多模型协作还能有效规避单点故障带来的风险，通过模型之间的互补与协同优化，提高系统整体的鲁棒性、响应速度以及服务质量，从而确保系统在复杂多变的需求环境中依然能够高效运行，满足用户的长期稳定需求。此外，模型互联网还具备良好的可扩展性，可以随着技术的发展和新模型的引入进行迭代升级，进一步强化协作能力，为各类应用场景提供持续可靠的解决方案。

## 3 未来研究方向

1）模型异构性的分析

模型互联网中包含各种各样的异构模型。模型间差异主要来源于训练数据分布、模型架构以及参数规模等因素。首先，不同模型在任务类型与专业领域上的性能存在显著差异。其次，模型之间存在形成内部集群的倾向。最后，不同模型在推理时延、推理成本以及上下文长度等效率与资源属性方面存在明显差异。如何定义模型之间的差异性对于匹配模型和任务至关重要。

现有研究通过在各领域的开源数据集上对模型按模型能力进行划分，例如 Song 等人[39]将能力定义为 25 个维度，然后根据各模型在不同任务上的表现进行人为判别打分。尽管能力区分模型在现有研究中被广泛应用并验证了其有效性，但是如何定义模型的异构性仍面临以下挑战：首先，业界对于如何划分统一的能力维度缺乏共识，难以形成有效的模型评估框架。此外，部分评估方法仍依赖人工干预，这不可避免地会引入主观偏差。最后，现有开源数据集的覆盖范围有限，难以全面覆盖所有专业领域。因此，如何定义模型异构是匹配任务的关键，缺乏异构性的共识已成为制约模型精准交互的重要瓶颈。

2）模型的协作方式

模型互联网集合了大量异构模型，所面临的任务类型丰富多样。因此，模型间的协作必须能够根据任务需求动态构建、实时适配。

现有研究多采用预设的固定模型与静态交互模式，主要聚焦于输出层面的集成策略，因而在动态任务场景下难以兼顾系统效率。例如，TETRIS 方法[40]依赖专家模型进行持续指导，导致关键模型资源被长期占用，从而降低了整体系统的运行效率。现有协作方法多采用静态且无差别的权重分配策略，未能根据任务情境与模型输出的动态效用进行适配性调整。因此，现有研究难以直接迁移到动态变化的模型互联网中。

模型的动态性、任务的多样性及网络状态的不确定性，共同构成一个复杂的动态环境，直接影响协作效能。因此，模型间的自适应协作是提升模型互联网性能的关键因素。

3）模型的动态管理

系统动态性是模型互联网的核心特征，它体现为模型节点能够自由地接入与退出。这种“即插即用”的特性带来了显著优势：当部分模型因故障、维护、网络障碍或升级而无法服务时，系统能够通过快速调度其他可用模型来维持服务，从而保障用户的体验质量不受影响。模型的分布式更新与替换也在

这一环境下得以高效实现。

然而，这种高度的动态性使得传统的静态配置与集中式管理范式完全失效。核心挑战在于：难以预设所有参与节点的状态，必须应对一个拓扑与能力节点实时演化的开放环境。A2A 方法[41]通过让智能体在加入网络时提交一份“Agent Card”来对模型进行标签化管理。这种方法为模型的发现与筛选提供了便利，但其本质上依赖于模型的“自我报告”机制。该方案在一个开放、动态且可能存在竞争或恶意的环境中暴露出 2 个致命缺陷：其一，虚假的“Agent Card”会污染整个模型生态，从而影响系统整体服务质量；其二，当某些模型因网络拥塞、硬件故障或被动下线而无法使用时，系统若不能即时感知并更新其可用状态，会严重造成资源浪费。

因此，高效且自适应的动态管理是模型互联网的运行基石。它既是实现模型“自由接入、即插即用”的前提，更是确保全局服务质量与系统可靠性的根本保障。

4）隐私与安全问题

模型互联网促进模型间的交互与协作，但单个模型节点的安全漏洞可能通过协作关系在系统扩散。被攻陷的模型节点可成为攻击者在系统内部的切入点，利用信任关系向其他智能体传播恶意指令或虚假信息，这种现象被称为“传染效应”[42]。例如，串行协作通过在提示词中嵌入恶意提示诱导模型执行未授权的指令，从而导致整个系统的全面妥协[43]。此外，现有研究发现大模型能够从用户提示中获取敏感信息[44]，而开放的模型互联网可能放大模型隐私泄露的风险。

要构建安全可信的模型互联生态环境，面临以下挑战：首先，如何甄别出恶意的模型从而在源头遏制恶意信息的传递；其次，如何保证交互时不泄露用户的敏感信息。面对上述挑战，未来的研究重心需从 2 个前沿展开攻坚：一方面，要发展出能实时识别并遏制恶意模型的动态防御体系；另一方面，必须攻克在复杂交互中保护数据隐私的关键技术，杜绝敏感信息泄露。这是迈向真正安全可信的模型互联网的必经之路。

5）异构模型互联互通

模型互联网在于构建一个泛在互联的模型网络群体，从而支持异构模型的高效传递和共享。然而，与互联网中的节点通信形式不同，模型互联网没有预先设定的通信链路，模型互联网的“连接”是瞬时、动态的软连接。此外，模型之间的组织形式也影响着模型路由的选择。

现有的多智能体通信协议在一定程度上为模型协作之间的通信提供了思路，但仍面临着挑战。例如，智能体通信协议（agent communication proposal，ACP）[45]是一种基于 RESTful API 的支持智能体通信开放标准。但在大规模部署的复杂场景下，该标准仍面临两大核心挑战：首先，缺乏动态协商、信息共享和模型协调的统一框架；其次，尚未建立跨域、可互认的信任体系以确保协作安全。

模型互联网旨在构建跨主体、跨平台的智能模型互联体系，实现模型之间的自由发现、自主协作与信息交换。因此，模型互联互通的研究不仅需要探讨模型间协作方式，还需要在协议层面建立通用、安全可信和高效的通信机制。

## 4 结束语

模型互联网是一种面向网络化人工智能模型的新范式。本文提出了模型互联网的概念、构想和层次结构，并搭建了相关原型系统。实验结果表明在精准度、稳定性、创造性和泛化性方面具有良好性能。模型互联网将探索模型间高效互联、智能互通、能力互补的新模式，以期推动大模型技术朝网络化方向发展，为人工智能领域研究提供新思路。此外，本文初步展望了模型互联网中富有潜力的研究方向，为深化与拓展模型互联网的研究提供了参考。

## 参 考 文 献

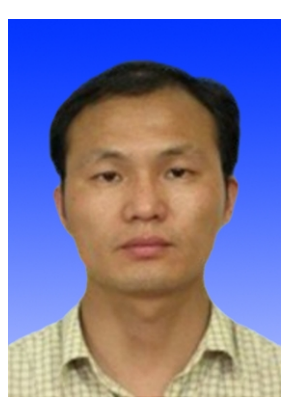

**Li Zhetao**, born in 1980. PhD, professor, PhD supervisor. Member of CCF. His main research interests include intelligent network and artificial intelligence.

**李哲涛**, 1980 年生。博士，教授，博士生导师。CCF 会员。主要研究方向为智能网络、人工智能。

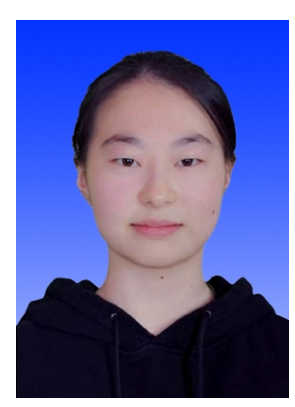

**Zeng Xiyu**, born in 1999. PhD candidate. Her main research interests include data privacy and artificial intelligence.

**曾曦玉**, 1999 年生。博士研究生。主要研究方向为数据隐私、人工智能。

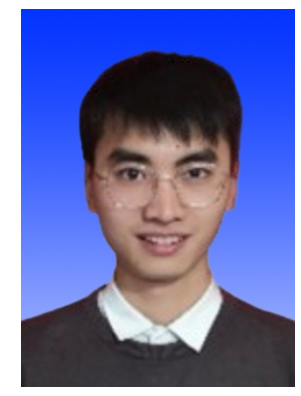

**Wang Jianhui**, born in 1997. PhD candidate. His main research interests include mobile edge computing and artificial intelligence.

**王建辉**, 1997 年生。博士研究生。主要研究方向为移动边缘计算、人工智能。

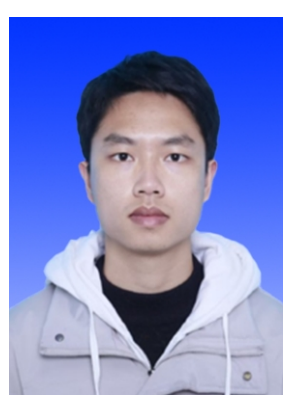

**Xiao Yong**, born in 2000. PhD candidate. His main research interests include mobile edge computing and artificial intelligence.

**肖　勇**, 2000 年生。博士研究生。主要研究方向为移动边缘计算、人工智能。

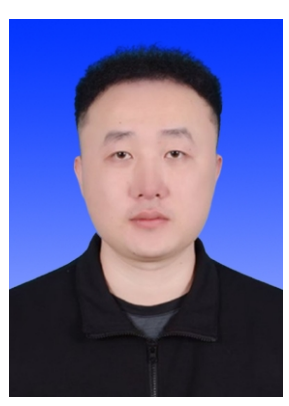

**Liu Zhongren**, born in 1996. PhD candidate. His main research interests include mobile edge computing and artificial intelligence.

**刘忠仁**, 1996 年生。博士研究生。主要研究方向为移动边缘计算、人工智能。

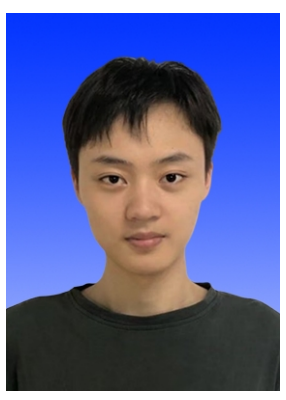

**Wu Junru**, born in 2000. PhD candidate. His main research interests include mobile edge computing and artificial intelligence.

**吴俊儒**, 2000 年生。博士研究生。主要研究方向为移动边缘计算、人工智能。

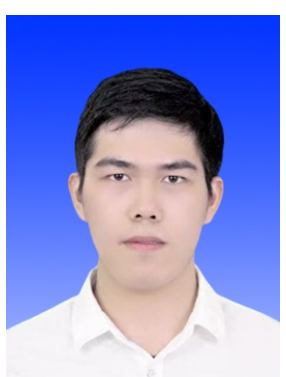

**Lai Junjie**, born in 2001. Master candidate. His main research interests include large language models and natural language processing.

**赖俊杰**, 2001 年生。硕士研究生。主要研究方向为大模型、自然语言处理。

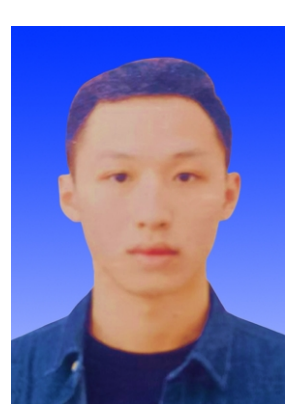

**Huang Jijun**, born in 2000. Master candidate. His main research interests include large language models and cybersecurity.

**黄纪俊**, 2000 年生。硕士研究生。主要研究方向为大模型、网络安全。

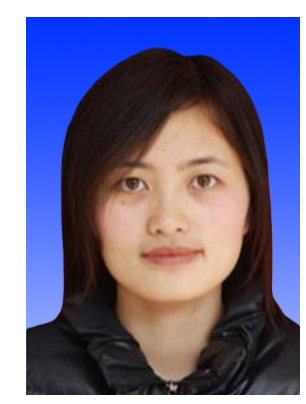

**Long Saiqin**, born in 1986. PhD, professor, PhD supervisor. Member of CCF. Her main research interest includes edge computing.

**龙赛琴**, 1986 年生。博士，教授，博士生导师。CCF 会员。主要研究方向为边缘计算。